\RequirePackage{fix-cm}
\documentclass[twocolumn]{svjour3}
\smartqed
\usepackage{bm}
\usepackage{graphicx}
\usepackage{newtxtext, newtxmath}
\usepackage{natbib}
\usepackage{balance}
\usepackage{amsmath}
\usepackage[citecolor=teal, colorlinks]{hyperref}
\usepackage{latexsym}
\usepackage[dvipsnames, svgnames, x11names]{xcolor}
\usepackage{makecell}
\usepackage{diagbox}
\usepackage{adjustbox}
\usepackage{caption}
\usepackage{subcaption}
\usepackage{soul}
\usepackage{booktabs}
\usepackage{multirow}
\usepackage{diagbox}
\usepackage{threeparttable}
\usepackage{colortbl}
\usepackage{tabularx}
\usepackage[figuresright]{rotating}
\usepackage{lipsum}
\usepackage{stfloats}
\usepackage{enumerate}
\usepackage{microtype}

\newcommand{\todo}[1]{\textcolor{magenta}{#1}}
\newcommand{\revise}[1]{\textcolor{black}{#1}}

\newcommand{\yfnote}[1]{\textcolor{cyan}{[YF: #1]}}

\definecolor{light-gray}{gray}{0.92}

\journalname{International Journal of Computer Vision}

\begin{document}

\title{A Comprehensive Study of the Robustness for LiDAR-based 3D Object Detectors against Adversarial Attacks}
\author{Yifan Zhang, Junhui Hou, Yixuan Yuan}
 

\institute{
	Yifan Zhang, Junhui Hou \at
	Department of Computer Science, City University of Hong Kong. \\
	\email{yzhang3362-c@my.cityu.edu.hk; jh.hou@cityu.edu.hk;}
        \and 
        Yixuan Yuan \at
        Department of Electronic Engineering, The Chinese University of Hong Kong. \\
        \email{yxyuan@ee.cuhk.edu.hk}
        \and 
This work was supported in part by the Hong Kong Research Grants
Council under Grant 11202320 and Grant 11219422, and in part by
the Hong Kong Innovation and Technology Fund under Grant
MHP/117/21.
}

\date{Received: date / Accepted: date}

\maketitle
\begin{abstract} 
Recent years have witnessed significant advancements in deep learning-based 3D object detection, leading to its widespread adoption in numerous applications. As 3D object detectors become increasingly crucial for security-critical tasks, it is imperative to understand their robustness against adversarial attacks. This paper presents the first comprehensive evaluation and analysis of the robustness of LiDAR-based 3D detectors under adversarial attacks. Specifically, we extend three distinct adversarial attacks to the 3D object detection task, benchmarking the robustness of state-of-the-art LiDAR-based 3D object detectors against attacks on the KITTI and Waymo datasets. We further analyze the relationship between robustness and detector properties. Additionally, we explore the transferability of cross-model, cross-task, and cross-data attacks. Thorough experiments on defensive strategies for 3D detectors are conducted, demonstrating that simple transformations like flipping provide little help in improving robustness when the applied transformation strategy is exposed to attackers. \revise{Finally, we propose balanced adversarial focal training, based on conventional adversarial training, to strike a balance between accuracy and robustness.} Our findings will facilitate investigations into understanding and defending against adversarial attacks on LiDAR-based 3D object detectors, thus advancing the field. \revise{The source code is publicly available at \url{https://github.com/Eaphan/Robust3DOD}.}

\keywords{3D object detection \and Point cloud \and Adversarial attack \and Robustness evaluation}
\end{abstract}

\section{Introduction} 
3D object detection, which aims to identify and localize the 3D bounding boxes of objects in specific classes, has been one of the most popular research fields in computer vision. 
The advancement driven by deep learning-based approaches~\citep{hu2022point,zhang2023upidet,zhang2023glenet,zhang2023spatial} broadens the applications of 3D object detection to diverse fields, such as autonomous driving, domestic robots, and virtual reality. 
While deep learning in 3D object detection has achieved revolutionary performance in recent years, the robustness of detectors has not been sufficiently investigated.

\begin{figure*}[htp]
	\centering
	\includegraphics[width=\textwidth]{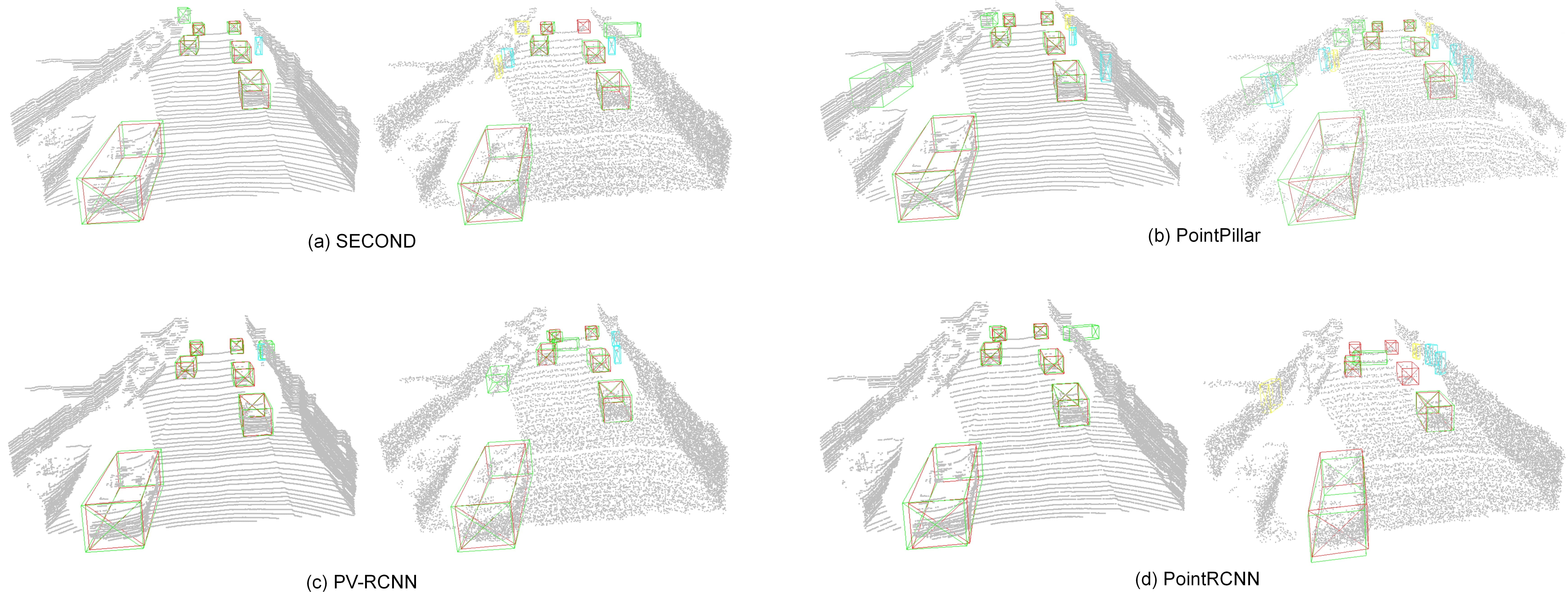} 
	\caption{
		Visual comparison of results predicted by various 3D detectors on (left) clean input and (right) under the PGD-based point perturbation attack ($\epsilon$=10cm). 
		The predicted bounding boxes of \textcolor{green}{car}, \textcolor{cyan}{pedestrian}, and \textcolor{yellow}{cyclist} are visualized in \textcolor{green}{green}, \textcolor{cyan}{cyan}, and \textcolor{yellow}{yellow}, respectively. The \textcolor{red}{ground-truth} bounding boxes are visualized in \textcolor{red}{red}. We recommend viewing the content in color and zooming in for better visibility of details.
  It can be observed that the various detectors output deteriorated predictions, i.e., more false positive boxes and not detected ground-truth boxes, under adversarial attacks.
	}
	\label{fig:introduction}
\end{figure*}

Existing studies show that deep neural networks are vulnerable to adversarial examples, which are intentionally generated to fool the network~\citep{szegedy2013intriguing}. The vulnerability may subsequently cause serious security problems and thus raise concerns about the deployment of neural networks in safety-critical applications.
We should also address potential security risks in 3D object detection applications, considering that the deteriorated predictions under attack (see Figure~\ref{fig:introduction}) will inevitably affect the reliability and stability of the system adopting 3D detectors. Therefore, the safety issues imply that the robustness of detectors against adversarial examples could be as important as the absolute accuracy on clean inputs.

To address the security issues, numerous attack~\citep{goodfellow2014explaining,MadryMSTV18} and defense~\citep{dziugaite2016study} methods have been proposed in the 2D image domain.
Besides, researchers have devoted significant effort to exploring attack and defense on learning-based 3D point cloud models~\citep{xiang2019generating,liu2022imperceptible,hu2023pointca}. 
Unfortunately, these existing works mainly concentrate on the task of point cloud classification, while the robustness of widely-used LiDAR-based 3D object detection, a significantly more complex task than point cloud classification, has not been systematically studied. Several works~\citep{TuRMLYDCU20,abdelfattah2021adversarial} generate adversarial examples by placing adversarial objects on the rooftop of target vehicles and successfully mislead 3D detectors, which can be attributed to the lack of relevant training data and mitigated by data augmentation.
In contrast, our work extends three prevalent and less constrained adversarial attacks to LiDAR-based 3D object detection: point perturbation, detachment, and attachment attacks. These attacks are \textit{plausible}, even without an active malicious agent, given the inevitable range errors of LiDAR sensors and other inherent factors.

\color{black}
This paper comprehensively investigates the robustness of LiDAR-based 3D detectors by (1) evaluating the effectiveness of different adversarial attacks for 3D object detectors, (2) analyzing the influence of \textit{network properties} on the robustness of detectors, and (3) exploring how to effectively \textit{defend} against adversarial attacks. 

\textit{First}, we rigorously evaluate the robustness of eight state-of-the-art 3D detectors against the adversarial attacks on two popular datasets, i.e., KITTI and Waymo~\citep{Sun_2020_CVPR} datasets.
Notably, the point perturbation attack is generally stronger than the other two attacks with a low perceptibility of its corresponding adversarial examples. 
Interestingly, the voxel-based methods show stronger robustness than the point-based detectors against the point perturbation attack.
Therefore, we discuss the influence of voxelization and voxel size on the robustness of voxel-based detectors. We identify the inherent disadvantages of the point perturbation attack when applied to voxel-based detectors, which make the attack less effective.
Besides, our comprehensive ablation studies unravel how the local feature integration in pillar encoding compromises the robustness of PointPillar against the point perturbation attack.
A deeper dive into the influences of model property is elaborated upon in our experiments section.

\textit{Second}, we provide an extensive evaluation of the effectiveness of the \textit{black-box} attack, i.e., the transferability of adversarial attacks including cross-model, cross-task, and cross-domain transferred attacks without available models or known ground-truth bounding boxes. 
For the \textit{cross-model} attack, adversarial examples exhibit some degree of transferability across various detectors. Adversarial examples generated based on vulnerable point-based detectors, like PointRCNN and 3DSSD, display impressive versatility in disrupting other detectors. 
For the \textit{cross-task} attack, the transfer perturbation attack is significantly more effective than the direct source attack on voxel-based detectors. This observation could help us design powerful and generalizable attacks from the perspective of the attacker.
For the \textit{cross-domain} attack, it does not exceed the effectiveness of source attacks in most cases. This observation may imply that the efficacy of adversarial attack techniques is somewhat tied to the data distribution of their originating domain.

\textit{Third}, we provide a thorough evaluation of various defense strategies against adversarial attacks.
We diagnose that the defense of simple transformation turns out to be of little help to improve the robustness assuming the defense strategy is publicly available knowledge for attackers.
On the other hand, adversarial training could improve the robustness of 3D detectors, but inadvertently compromise the accuracy of models on clean examples. To this end, we propose Balanced Adversarial Focal Training (BAFT) to harmonize the losses on clean and adversarial examples during training and finally achieve a better balance between robustness and accuracy.

In summary, the main contributions of this paper are:
\begin{itemize}
	\item[$\bullet$] we extend three kinds of adversarial attacks to LiDAR-based 3D object detection, including point perturbation, detachment, and attachment;
	
	\item[$\bullet$] 
	we benchmark the robustness of various detectors against the adversarial attacks on KITTI and Waymo datasets, and analyze the influence of feature representations, architectures, model sizes, etc;
	
	
	
	\item[$\bullet$] we provide an extensive evaluation of the effectiveness of the black-box attack, i.e., the transferability of adversarial attacks including cross-model, cross-task, and cross-domain transferred attacks;
	
	\item[$\bullet$] we provide a thorough evaluation of various defense strategies against adversarial attacks and conclude that the defense of simple transformation turns out to be of little help to improve the robustness assuming the defense strategy is publicly available knowledge for attackers;
	
	\item[$\bullet$] we propose Balanced Adversarial Focal Training (BAFT) to harmonize the losses on clean and adversarial examples.
\end{itemize}

\color{black}

To the best of our knowledge, this represents the first comprehensive evaluation and rigorous analysis of the robustness of LiDAR-based 3D object detectors. We believe that our findings will drive future endeavors to enhance the robustness and reliability of 3D object detectors against adversarial attacks without sacrificing detection accuracy.
The remainder of the paper is structured as follows: Section~\ref{sec:related_work} reviews the existing literature on LiDAR-based detectors and adversarial attacks. In Section~\ref{sec:adversarial_attacks}, we adapt three adversarial attacks to 3D object detection. Then, in Section~\ref{sec:defense_method}, we introduce nine general defense strategies as well as our BAFT approach. Section~\ref{sec:experimental_setup} details our experimental setup. Section~\ref{sec:evaluation_of_adversarial_attacks} provides an experimental evaluation and analysis of the robustness of state-of-the-art models against different adversarial attacks, while Section~\ref{sec:transferability_evaluation} focuses on the black-box attacks. In Section~\ref{sec:eval_defense}, we assess the efficacy of commonly used defense strategies for point cloud tasks. Section~\ref{sec:eval_baft} validates the superiority of our proposed BAFT over traditional adversarial training. Finally, Section~\ref{sec:conclusion} summarizes and concludes our study.


\section{Related Work}\label{sec:related_work}
\subsection{LiDAR-based 3D Object Detection}
We can roughly classify 3D object detection into two groups. 1) \textit{Voxel-based 3D detectors} typically turn point clouds into grid-structure forms with fixed sizes. Early work VoxelNet~\citep{Zhou_2018_CVPR} first proposed to convert the point cloud to a regular volumetric representation by stacked voxel feature encoding. Extracting features with a conventional convolution network is inefficient due to the sparsity of non-empty voxels, so \cite{yan2018second} introduced a more efficient sparse convolution to accelerate training and inference. \cite{Lang_2019_CVPR} deployed a novel encoder that learns features on pillars (vertical columns) of the point clouds, where only 2D convolutional layers are used. \cite{He_2020_CVPR} employed auxiliary tasks including center estimation and foreground segmentation to guide the network to learn the intra-object relationship. 
\cite{deng2021voxel} proposed voxel RoI pooling to directly extract RoI features from coarse voxels.
\textit{2) Point-based 3D detectors}
consume the raw 3D point clouds directly and generate predictions based on (down-sampled) points. 
\cite{shi2019pointrcnn} presented a strong point-based feature extractor and generated high-quality proposals on foreground points. 3DSSD~\citep{yang20203dssd} introduced a new sampling strategy as a supplement to D-FPS, namely F-FPS, to preserve more foreground points within target objects. It also built an anchor-free regression head to predict 3D bounding boxes based on representative points. ~\cite{zhang2022not} and~\cite{chen2022sasa} optimized the down-sampling strategies with semantic information to preserve foreground points as much as possible.
\cite{shi2020points} proposed to encode the point cloud efficiently in a fixed-radius near-neighbor graph, and predict boxes of object that each vertex in the graph belongs to.
Besides, PV-RCNN~\citep{shi2020pv} integrated the multi-scale voxel-based feature representation and point-based features containing accurate location information.
While most existing works focus on improving the detection accuracy and efficiency of detectors, our study emphasizes their robustness against adversarial attacks.

\subsection{Adversarial Attacks on 2D Images}
~\cite{szegedy2013intriguing} first pointed out that the neural network is susceptible to carefully designed adversarial perturbations. 
An adversarial example that is similar to the original data can successfully fool the neural networks to output wrong predictions. The community has expressed concern about the vulnerability of deep learning models and many works have been proposed to search for effective adversarial attacks and possible defense strategies.
~\cite{goodfellow2014explaining} proposed the fast gradient method, which uses the gradient calculated by the loss with respect to the image.
~\cite{kurakin2018adversarial} further extended FGM to an iterative approach (I-FGM), which leads to a higher success rate of attack than basic FGM.
In order to stabilize update directions and escape from local maxima, MI-FGM~\citep{dong2018boosting} introduced a momentum term into the iterative process. The projected gradient descent (PGD)~\citep{MadryMSTV18} is similar to MI-FGM except for random initialization of perturbation at the beginning of the iteration.
~\cite{carlini2017towards} proposed the optimization-based C\&W attack that finds the closest adversarial example to the original example based on three norm distances. Besides, there are some studies of physically realizable adversarial examples instead of only dealing with collected 2D image data~\citep{hu2022adversarial,cheng2022physical}.\\

\noindent\textbf{Defense on Adversarial Attack.} Against adversarial attacks, numerous defense methods have been proposed.
Adversarial training~\citep{goodfellow2014explaining,Herrmann_2022_CVPR} is a widely investigated and effective defense method. It involves training a model from scratch using data augmented with adversarial examples. Despite the improvement in tasks of the image domain, adversarial training is more time-consuming than training on clean data and takes more epochs to fit adversarial examples~\citep{tramer2018ensemble}.
Another defense approach involves transforming adversarial examples before feeding them to the models.
~\cite{dziugaite2016study} and~\cite{guo2018countering} proposed to preprocess images with JPEG compression to mitigate the effect of adversarial perturbation. 
~\cite{xu2017feature} enhance the robustness of neural networks by reducing the color bit depth of each pixel and spatial smoothing.
~\cite{osadchy2017no} employed a set of filters, such as the median filter, average filter, and Gaussian low-pass filter, to remove the adversarial noise.
\cite{li2019certified} added random noise to pixels of adversarial examples to eliminate the effects of adversarial perturbations.

\subsection{Adversarial Attacks on 3D Point Clouds}
Following previous studies in the image domain, numerous works on adversarial attacks and robustness have been proposed in the 3D vision field. Existing 3D adversarial attacks can be roughly classified into three categories: shifting points~\citep{xiang2019generating}, adding new points~\citep{liu2020adversarial}, dropping points~\citep{zheng2019pointcloud,wicker2019robustness}.
~\cite{xiang2019generating} first proposed 3D adversarial attacks, including adversarial point perturbation, adversarial point generation and adversarial objects generation, against 3D learning models and show effectiveness in attacking PointNet~\citep{qi2017pointnet}.
~\cite{wen2020geometry} further considered the local consistency of curvatures between the surface of clean input and adversarial examples.
 \cite{liu2019extending} extended FGSM method to generating 3D adversarial examples using triangular mesh constrained with $L_{\infty}$ norm.
~\cite{tsai2020robust} applied C\&W attack~\citep{carlini2017towards} with incorporated KNN distance as extra constraint on perturbation.
\cite{hamdi2020advpc} utilized auto-encoders with a novel data adversarial loss to generate network-transferable perturbations.
To maintain geometric smoothness, ~\cite{liu2022imperceptible} suggested limiting point-wise perturbations to follow the normal direction within a strictly defined width.
~\cite{liu2020adversarial} verified adding simple features into the original point clouds can generate adversarial point clouds.
~\cite{zheng2019pointcloud} constructed a saliency map and realized attacking models by dropping out the critical points.
Typical defense methods in point cloud fields include simple random sampling~\citep{jia2021intrinsic}, upsampling after denoising~\citep{zhou2019dup}, gather vector~\citep{dong2020self}, provable defense~\citep{Liu_2021_CVPR}, adversarial training~\citep{sun2021adversarially,liu2019extending}.\\

\noindent\textit{Adversarial Attacks on LiDAR-based Detectors.}
Studies for adversarial attacks on 3D object detection have been scarce so far. 
~\cite{TuRMLYDCU20} and ~\cite{abdelfattah2021adversarial} generated adversarial examples by placing adversarial objects on the rooftop of target vehicles, the successful attack of which can be attributed to the lack of relevant training data and mitigated by data augmentation. ~\cite{tu2021exploring} learned an adversarial textured mesh in a differentiable manner to attack any multi-sensor 3D detectors by rendering the adversarial mesh into both LiDAR and image inputs.
~\cite{sun2020towards} constructed a class of spoofing attacks by injecting a small number of spoofed LiDAR points and achieved high success rates on target models. ~\cite{lehner20223d} proposed to achieve stronger domain generalization ability of detectors by a novel data augmentation that adversarially deforms point cloud of objects along the ray.



\section{Adversarial Attacks}\label{sec:adversarial_attacks}

In this section, we delineate the methods used to produce adversarial examples that target LiDAR-based 3D detectors.
Let $\mathbf{X}=\{\bm{p}_{i}|i=1,2,...,N\}$ represent the input point cloud, where $N$ indicates the number of points, and each point $\bm{p}_{i}\in\mathbb{R}^{3}$ is characterized by its 3D coordinates $(x_p,y_p,z_p)$. We temporarily disregard other point properties like intensity and elongation.
The ground-truth 3D bounding boxes for point cloud $\mathbf{X}$ are denoted by $\mathcal{G}=\{\mathbf{B}_j\}$, where ${\mathbf{B}_j}=(x_j,y_j,z_j,w_j,l_j,h_j,r_j,c_j)\in\mathbb{R}^8$. Here, $(w_j,l_j,h_j)$ defines the size of the box, $(x_j,y_j,z_j)$ specifies the coordinates of the box center, and $r_j$ represents the orientation angle.

The goal of adversarial attacks is to subtly modify points in $\mathbf{X}$ to produce an adversarial example, $\mathbf{X}_{a}$, in such a way that deep detectors are deceived into producing incorrect detection results.
These adversarial examples can be feasible in reality and remain \textit{plausible} even in the absence of a malicious attacker.
This is because real-world scenarios might involve inevitable range error of LiDAR sensors, diminished laser reflectivity on dark objects, and adverse weather conditions, among other factors.
We have adapted three types of adversarial attack methods, commonly investigated in the point cloud domain~\citep{xiang2019generating,zheng2019pointcloud}, to suit LiDAR-based 3D object detection.
For comprehensiveness, we elaborate on the generation algorithms of adversarial examples targeting 3D object detection in the subsequent portions of this section. Particularly, we discuss the limitation of point perturbation attacks when applied to voxel-based detectors.

\subsection{Adversarial Point Perturbation}\label{sec:adversarial_point_perturbation}
In perturbation attacks, we derive the adversarial example $\mathbf{X}_{a}$ by shifting the 3D coordinates of raw points. 
Specifically, the adversarial perturbation of coordinates can be represented as $\delta=\{\delta_i\in \mathbb{R}^3|i=1,...,N\}$, and thus, $\mathbf{X}_{a}=\{\bm{p}'_i:=\bm{p}_i+\delta_i|i=1,...,N\}$.
We adopt the non-targeted attack where the adversary does not have a specific target of bounding boxes but just manipulates the prediction of detectors on $\mathbf{X}_a$ to be different from the ground-truth labels.
The problem can be further formulated as a gradient-based optimization algorithm:
\begin{equation}\label{eq:gradient_optimize}
	\mathop{\min}  J(\mathbf{X}_a, \mathcal{G}) = -L_{det}(\mathbf{X}_a, \mathcal{G}) + \lambda \mathcal{D}(\mathbf{X}, \mathbf{X}_a),
\end{equation}
where $L_{det}(\cdot)$ is the detection loss,
 $\mathcal{D}(\cdot)$ is the distance function measuring the magnitude of perturbation $\delta$, 
and the non-negative parameter $\lambda$ balances the two terms. 
As the correspondence between $\mathbf{X}$ and $\mathbf{X}_a$ is known, we simply use the $\ell_2$ norm as the distance metric, denoted as $\mathcal{D}_{L2}$, to restrain the magnitude of perturbations.
Then we can get the gradient of $J$ with respect to $\mathbf{X}_a$ via backpropagation:
\begin{equation}\label{eq:calculate_gradient}
	\bigtriangledown_{\mathbf{X}_a} J(\mathbf{X}_a, \mathcal{G})  = \frac{ \partial(J(\mathbf{X}_a, \mathcal{G}))}{\partial \mathbf{X}_a},
\end{equation}
Taking the iterative fast gradient method (I-FGM)~\citep{kurakin2018adversarial} as an example, we can obtain the adversarial example as:

\begin{equation}\label{eq:adversarial_attack_iteration} 
	\mathbf{X}_a^0=\mathbf{X},~~\mathbf{X}_a^{t+1} = \texttt{Clip}_{\mathbf{X},\epsilon}\left\{  \mathbf{X}_a^t - \alpha \frac{ \bigtriangledown_{\mathbf{X}_a^t} J(\mathbf{X}_a^t, \mathcal{G}) }{\|  \bigtriangledown_{\mathbf{X}_a^t} J(\mathbf{X}_a^t, \mathcal{G})  \|_2}\right \}, 
\end{equation}
where $\mathbf{X}_a^t$ denotes the adversarial example after $t$ iterations, $\alpha$ is the step size, and $\texttt{Clip}_{\mathbf{X},\epsilon}\{\hat{\mathbf{X}}\}$ is the function performing point-wise clipping of point cloud $\hat{\mathbf{X}}$ (an intermediate variable), defined as:
\begin{equation}\label{eq:perturbation_clip}
	\texttt{Clip}_{\mathbf{X},\epsilon}\{\hat{\mathbf{X}}\} = \{\tilde{\bm{p}}_i| \tilde{\bm{p}}_i = \texttt{clip}(\hat{\bm{p}}_i), i=1,...,N\},
\end{equation}

\begin{equation}
	\texttt{clip}(\hat{\bm{p}_i}) =
	\begin{cases}
		\hat{\bm{p}}_i, & |\hat{\bm{p}}_i - \bm{p}_i|\le \epsilon \\
		\bm{p}_i+\epsilon\frac{|\hat{\bm{p}}_i-\bm{p}_i|}{|\hat{\bm{p}}_i-\bm{p}_i|}, & |\hat{\bm{p}}_i-\bm{p}_i|> \epsilon
	\end{cases}
\end{equation}
so the maximum shift distance of each point ($\ell_2$ norm of the perturbation) is limited by parameter $\epsilon$.
The I-FGM update rule can also be replaced with MI-FGM~\citep{dong2018boosting}, PGD~\citep{MadryMSTV18} or degraded to FGM~\citep{goodfellow2014explaining}.\\

\noindent\textbf{Attacks on Voxel-based Detectors.} Different from point-based detectors directly consuming raw point clouds, voxel-based methods first convert the point cloud to a regular volumetric representation before applying 3D convolution. Most implementations of state-of-the-art methods initialize the voxel features with mean voxel feature extraction (mean-VFE): denote $\mathbf{V}=\{p_i|i=1,2,...,N_\mathbf{V}\}$ as a non-empty voxel containing $N_\mathbf{V}$ points, the feature of voxel initialized via:
\begin{equation}
	f_\mathbf{V}=\frac{1}{N_\mathbf{V}} \sum_{p_i\in \mathbf{V}}p_i,
\end{equation}
then $f_\mathbf{V}$ is commonly fed in a sparse convolution network. 
However, due to the rule of backpropagation, points in the voxel $\mathbf{V}$ share identical gradients during adversarial attacks, leading to the same perturbations for each point. This homogeneous perturbation diminishes the effectiveness of attacks, as it does not allow for unique perturbations tailored to individual points.

Moreover, though gradient-based adversarial attacks have shown generalization on various tasks in the image domain, they may encounter a \textit{dilemma} when employed in voxel-based detectors. 
While the location of the grid is fixed and only the value of the pixel is perturbed when attacking an image-based task, the voxel that the point belongs to may change after the point perturbation attack since the coordinates of the point change. 
If we perform re-voxelization on the perturbed point cloud, the deliberately disturbed perturbation on points may not work for new voxel representation.
On the other hand, the sparse 3D convolution is performed on voxel-based representation based on not only the value of voxels but also the location of voxels~\citep{yan2018second}.
If we do not conduct re-voxelization, the \textit{location} of voxel $\mathbf{V}$, reflecting the geometric information of the object and can be used as a basis for detection, is not affected by attacks though the \textit{value} of $\mathbf{V}$ is perturbed.
These complications amplify when the voxel size is set smaller, since the perturbed point is more likely to move beyond the original voxel it belongs to. We evaluate the effectiveness of point perturbation attacks against voxel-based detectors in Section~\ref{sec:attacking_voxel_based_detectors}.

\begin{figure*}[htp]
	\centering
	\includegraphics[width=\textwidth]{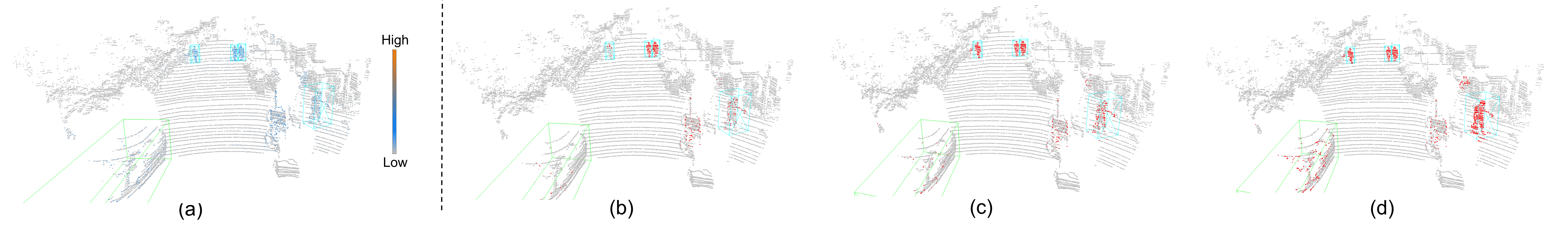} 
	\caption{
		(a) A saliency map based on the gradient of the loss with respect to the input image. The color of the point represents the size of its gradient. Besides, the GT boxes of \textcolor{green}{cars} and \textcolor{cyan}{pedestrians} are shown in \textcolor{green}{green} and \textcolor{cyan}{cyan}. (b-d) The illustration of \textcolor{red}{detached points} (visualized in \textcolor{red}{red}) at the 3rd, 7th, and 10th iterations, respectively (removing 2\% of the point cloud with a total of 10 iterations). Best viewed in color and zoom in for more details.
	}
	\label{fig:vis_detachment_illustration}
\end{figure*}

\subsection{Adversarial Point Detachment}
In practical scenarios, the obtained point clouds may be sparser due to occlusion, low-reflectivity objects, and extreme weather. Hence, we also present an adversarial setting by removing \textit{critical} points in a similar way to ~\cite{zheng2019pointcloud}.
For given point clouds, the magnitude of point-wise gradients reflects the influence of local position disturbance on the detection loss/predictions. 
Therefore we construct a saliency map with point-wise gradients derived by backpropagation, and remove points according to the saliency map.
Specifically, the importance of each point is measured by the sum-of-square of the gradients of the loss with respect to the coordinates:
\begin{equation}\label{eq:saliency_score}
	s_i = \| \partial L_{det}/\partial \bm{p}_{i} \|_2.
\end{equation}
Then saliency scores $\{s_i\}_{i=1}^N$ of all points form a saliency map.
Considering the dependencies between points, we drop the points \textit{iteratively} such the points already removed will not be involved in the current iteration. 
For a better understanding of the adversarial removed points, we visualize the saliency map and the point detachment process in Fig~\ref{fig:vis_detachment_illustration}.



\subsection{Adversarial Point Attachment}
The point attachment-based adversarial attack is implemented by adding new points $\mathbf{Z}=\{z_i\in \mathbb{R}^3|i=1,...,M\}$ at appropriate locations to $\mathbf{X}$. To tackle the challenge posed by the large search space of adding points,  following~\cite{xiang2019generating}, we adopt an \textit{initialize-and-shift} method. 
First, we obtain a saliency map via Eq.~\eqref{eq:saliency_score} and consider the region where saliency points are located to be vulnerable. Accordingly, we choose the points with top-$K$ saliency scores as initial points.
Then we shift the initial points, which are further incorporated into $\mathbf{X}$ to form an adversarial example, i.e., $\mathbf{X}_a=\mathbf{X} \cup \mathbf{Z}$. The appropriate displacement of initial points can be obtained by solving the optimization problem formulated in Eq.~\eqref{eq:gradient_optimize}. The generation of the adversarial examples follows an iterative training process, guided by the detection loss and constraint of imperceptibility.
Particularly, we adopt the Chamfer distance $\mathcal{D}_{C}$ as the constraint between $\mathbf{X}_a$ and $\mathbf{X}$:
\begin{equation}\label{eq:chamfer_distance}
	\begin{aligned}
		\mathcal{D}_{C}(\mathbf{X}, \mathbf{X}_a) = & \frac{1}{\left\| \mathbf{X} \right\|_0 } \sum_{p\in \mathbf{X}} \min_{p'\in \mathbf{X}_a }  \left\| p - p' \right\| \\
		& + \frac{1}{\left\| \mathbf{X}_a \right\|_0 } \sum_{p'\in \mathbf{X}_a} \min_{p\in \mathbf{X} }  \left\| p - p' \right\|.
	\end{aligned}
\end{equation}
Note that only the added points are updated in each iteration described in Eq.~\eqref{eq:adversarial_attack_iteration}.

\color{black}
\subsection{Real-world Plausibility of Adversarial Examples}\label{sec:feasibility}
In this part, we elucidate the real-world plausibility of the above three adversarial attack paradigms discussed.\\


\noindent\textbf{Point Perturbation Attack.} Real-world LiDAR measurements are prone to coordinate inaccuracies due to several intrinsic and extrinsic factors:
\begin{itemize}
	\item[$\bullet$] \textit{Sensor System Limitations}: Every LiDAR sensor, owing to its mechanical and optical construction, has inherent susceptibilities that can introduce deviations in both distance and angular measurements.
	\item[$\bullet$] \textit{Environmental Interferences}: Natural environmental conditions such as rain, snow, and atmospheric disturbances can attenuate and scatter the laser beams. This not only weakens the returning signal but can also skew distance calculations, thereby introducing measurement inaccuracies.
	\item[$\bullet$] \textit{Motion-induced Distortions}: For a LiDAR mounted on fast-moving platforms, the discrete time sampling coupled with continuous motion can result in distortions in the captured point cloud, leading to coordinate discrepancies.
\end{itemize}

\noindent\textbf{Point Detachment Attack.} Several factors can lead to incomplete or omitted point captures in LiDAR scans:
\begin{itemize}
	\item[$\bullet$] \textit{Surface Reflectivity Challenges}: Surfaces with suboptimal reflectivity, due to material properties or incident angles, can lead to the inefficient return of LiDAR beams, causing missing data points.
	\item[$\bullet$] \textit{Weather-induced Absorption}: Extreme weather conditions such as dense fog, heavy rain, or dust storms can lead to significant absorption of LiDAR pulses, resulting in missing points in the scan.
	\item[$\bullet$] \textit{Sensor Anomalies}: External factors like accumulated dust or insect agglomerates on the surface of LiDAR can occasionally interfere with beam emission, leading to missing data points.
\end{itemize}

\noindent\textbf{Point Attachment Attack.} The crowded nature of modern roads means that multiple LiDAR-equipped vehicles often operate in close proximity. This can result in the unintentional capturing of light pulses from neighboring LiDAR systems, especially if they operate at similar frequencies~\citep{brinon2021methodology}. Such interference, commonly referred to as crosstalk, often introduces extraneous data points in the point cloud.

\color{black}

\section{Defense Mechanisms}\label{sec:defense_method}
\subsection{General Defense Strategy}
Since adversarial examples are produced by introducing unnatural patterns into point clouds, we can apply transformations that remove such patterns to defend the 3D detectors from attacks. In this paper, we study the following 9 defense methods and verify their effectiveness on the adversarial attacks towards 3D object detectors in Section~\ref{sec:eval_defense}.
\begin{itemize}
    \item[$\bullet$] \textbf{Adversarial Training.} It is one of the most studied defenses against adversarial attacks~\citep{goodfellow2014explaining}, improving the robustness of models against specific attacks by augmenting the training set with corresponding adversarial examples.

    \item[$\bullet$] \textbf{Simple Transformation}. We apply flipping, scaling, and rotation to point clouds as defense methods. The final detection results are derived by applying a contrary transformation to the output of models.
    
    \item[$\bullet$] \textbf{Gaussian Noise.} The location of each point is perturbed by values randomly drawn from a Gaussian distribution. 
    
    \item[$\bullet$] \textbf{Quantification.} The coordinates of point cloud data in centimeters are quantized as integers that can mitigate the small variations in the coordinates of points.
    
   \item[$\bullet$] \textbf{Data Augmentation.} We study the effectiveness of part-aware data augmentation (PA-AUG)~\citep{choi2021part} for improving the robustness of detectors against adversarial attacks.
    
  \item[$\bullet$] \textbf{Random Sampling.} We can obtain a subset of points where each point is chosen randomly with the same probability.
    
  \item[$\bullet$] \textbf{KNN-based Denoising.} The point clouds are first categorized into several clusters. According to the sigma rule, the point is considered an outlier and removed if it is not within 3 times the standard deviation of distance distribution from the center of cluster~\citep{carrilho2018statistical}.

\end{itemize}


%

\color{black}
\subsection{Balanced Adversarial Focal Training}

While adversarial training serves as an effective defense mechanism, it may inadvertently compromise the accuracy of models on clean examples. In this work, we train the detectors on both clean and adversarial samples. Directly summing these two losses, however, can lead to one overshadowing the other, resulting in sub-optimal performance on specific samples. Addressing this concern, we introduce a novel strategy termed \textbf{Balanced Adversarial Focal Training (BAFT)}, designed to balance the losses during training.

Initially, we compute the loss for clean examples, $L_{\text{clean}}$. Subsequently, adversarial examples $\mathbf{X}_a$ are derived, upon which the loss $L_{\text{adv}}$ is computed. Our method aims to harmonize these distinct losses.

To achieve this balance, we commence by normalizing the losses using the softmax function:
\begin{equation}
	p_{\text{clean}}, p_{\text{adv}} = \text{softmax}\left([- L_{\text{clean}}, - L_{\text{adv}}\right]).
\end{equation}
A higher value of either $p_{\text{clean}}$ or $p_{\text{adv}}$ indicates the relative ease of optimizing the corresponding examples.

Drawing inspiration from focal loss~\cite{lin2017focal}, we introduce weights to make detectors pay more attention to those hard-recognized examples. Specifically, the focal weights for clean and adversarial losses are defined as:
\begin{equation}
	\tilde{w}_{\text{clean}} = - (1 - p_{\text{clean}})^{\gamma} \log(p_{\text{clean}}) + \zeta,
\end{equation}
\begin{equation}
	\tilde{w}_{\text{adv}} = - (1 - p_{\text{adv}})^{\gamma} \log(p_{\text{adv}}) + \zeta,
\end{equation}
where $\gamma$ is a predetermined parameter, and $\zeta$ acts as a regularization term modulating the weights.

The normalized weights are then computed:
\begin{equation}
		w_{\text{clean}} = \frac{\tilde{w}_{\text{clean}}}{\tilde{w}_{\text{clean}} + \tilde{w}_{\text{adv}}},
\end{equation}
\begin{equation}
		w_{\text{adv}} = \frac{\tilde{w}_{\text{adv}}}{\tilde{w}_{\text{clean}} + \tilde{w}_{\text{adv}}}.
\end{equation}
If the model predicts wrong results on clean examples, it will lead to a diminished $p_{\text{clean}}$ and subsequently larger $w_{\text{clean}}$ for the clean examples.

By integrating these focal weights, the aggregated loss is formulated as:
\begin{align}
	L_{total} = w_{\text{clean}} \times L_{\text{clean}} + w_{\text{adv}} \times L_{\text{adv}}.
\end{align}

\color{black}

\section{Experimental Setup}\label{sec:experimental_setup}
\subsection{Datasets}\label{sec:dataset}

The \textbf{KITTI} dataset is one of the most commonly used datasets in the outdoor 3D object detection task. 
\revise{KITTI dataset adopts a single Velodyne HDL-64E sensor that produces a 360° view with 64 scan lines. The range of the sensor is approximately up to 120m for most objects.}
Specifically, the KITTI dataset consists of 7481 training samples with annotation of objects in the camera field of vision and 7518 testing samples. The input contains two modalities: point cloud and image. Following prior works~\citep{shi2020pv}, we further divide all training samples into two subsets with 3712 and 3769 samples for training and validation, respectively. The benchmark further divides the samples into three difficulty levels: easy, moderate, and hard based on the occlusion level, truncation ratio, and object size.\\ 

\begin{figure}[t]
	\centering
	\includegraphics[width=0.49\textwidth]{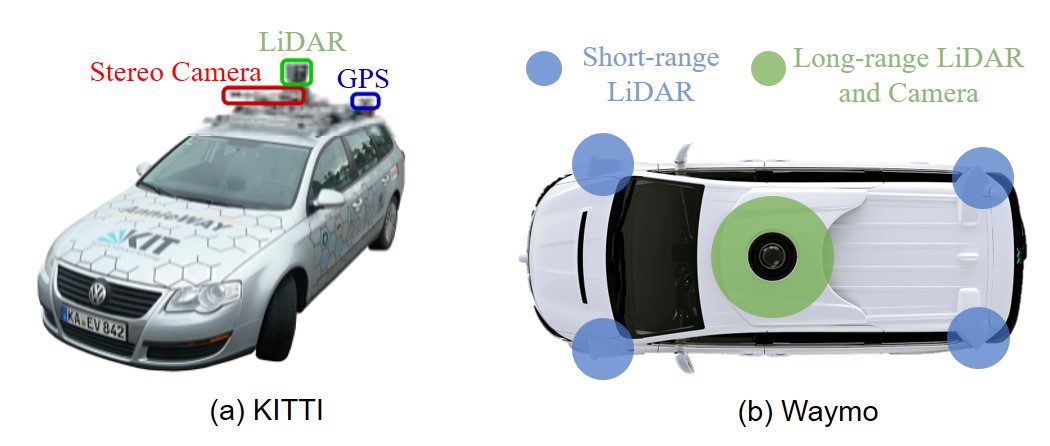} 
	\caption{
		\revise{Illustration of LiDAR sensor of KITTI~\cite{Geiger_KITTI} and Waymo dataset~\cite{Sun_2020_CVPR}.}
	}
	\label{fig:dataset}
\end{figure}

\noindent The \textbf{Waymo Open} Dataset~\citep{Sun_2020_CVPR} is a large-scale 3D object detection dataset for autonomous driving. 
\revise{The dataset uses multiple custom LiDAR sensors, including a long-range LiDAR and four short-rage LiDARs (see Fig.~\ref{fig:dataset}). It provides the first two returns of each laser pulse. Due to the combination of multiple LiDARs and the availability of the first two returns of each pulse, the resolution of points collected in the Waymo dataset is higher than that of the KITTI dataset.}
The dataset is collected in diverse scenes and contains 798 sequences (158361 LiDAR frames) for training, 202 sequences (40077 LiDAR frames) for validation, and 150 testing sequences. For each sequence, there are about 200 frames with LiDAR points, multi-view camera images and object annotations in full $360^\circ$. The objects are further categorized into two difficulty levels: LEVEL\_L1 for objects with more than five observed LiDAR points, and LEVEL\_L2 for objects with 1-5 points. The mAP and mAPH (mAP weighted by heading accuracy) are commonly adopted as metrics to evaluate the 3D object detectors.

\setlength{\tabcolsep}{2pt}
\begin{table}[htp]
	\centering
	\caption{Properties of the LiDAR-based 3D object detectors.}
	\label{table:detector_properties}
	\scalebox{0.85}{
		\begin{tabular}{l|ccc|cc|c} 
			\toprule
			\multirow{2}{*}{Detectors} & \multicolumn{3}{c|}{Representation} & \multicolumn{2}{c|}{Architecture} & \multirow{2}{*}{Params (M)}  \\
			& Point & Voxel & Point-Voxel        & One-stage & Two-stage     &                  \\ 
			\hline
			SECOND  &       & \checkmark    & & \checkmark        &               & 5.33             \\
			Voxel-RCNN                 &       & \checkmark    & &           & \checkmark            &   7.60        \\
			PDV     &       & \checkmark    & &           & \checkmark            & 12.86            \\
			PointPillar                &       & \checkmark    & & \checkmark        &               &    4.83   \\
			Part-A\^{}2  &       & \checkmark    & &           & \checkmark            &       63.81    \\
			PV-RCNN &       &       & \checkmark                 &           & \checkmark            &   13.12  \\
			3DSSD   & \checkmark    &       & & \checkmark        &               &    2.61   \\
			PointRCNN                  & \checkmark    &       & &           & \checkmark            &  4.04      \\
			\bottomrule
		\end{tabular}
	}
\end{table}

\subsection{Target LiDAR-based Detectors}
We consider eight state-of-the-art deep learning-based LiDAR detectors with different structures and properties, including SECOND~\citep{yan2018second}, Voxel-RCNN~\citep{deng2021voxel}, PDV~\citep{hu2022point}, PointPillar~\citep{LangVCZYB19}, Part-A\^{}2~\citep{shi2020points}, PV-RCNN~\citep{shi2020pv}, 3DSSD~\citep{yang20203dssd}, PointRCNN~\citep{shi2019pointrcnn}. Table~\ref{table:detector_properties} shows their characteristics in terms of the feature representation, network architecture, and the number of model parameters. The goal of all models is to detect 3 types of objects: cars, pedestrians, and cyclists. We use the publicly available code and pre-trained models as much as possible. For methods without available pre-trained models, we retrain the models with the default configuration provided by the original authors.

\subsection{Implementation Details}\label{sec:implementation_details}
\noindent\textbf{Adversarial Attacks.} First, for the point perturbation attack, we obtain the adversarial examples with FGM, MI-FGM, and PGD described in Section~\ref{sec:adversarial_attacks}. For the iterative attacks, the number of iterations is set to 40, and the step size $\alpha$ is set to $\epsilon / 30$. We evaluate the point perturbation attack when $l_2$ norm of perturbation $\epsilon$ is set to each value in \{0.5, 1, 3, 5, 7, 10\}cm. \textit{Second}, we evaluate the point detachment attack when setting the ratio of detached points to each value from \{0.001, 0.003, 0.005, 0.007, 0.01, 0.02\}, and the points are gradually detached in a total of 10 iterations. Thirdly, 164 and 500 points are attached to the raw point clouds in point attachment attack for KITTI and Waymo datasets respectively, while the $l_2$ norm of perturbation $\epsilon$ on \textit{attached} points take values in \{0.05, 0.1, 0.3, 0.5, 0.7, 1\}m. Our adversarial attack methods are implemented with Pytorch framework~\citep{Pytorch}.\\ 


\noindent\textbf{Robustness Measurement.} 
As the mAP value of different 3D detectors varies, we adopt the relative metric called mAP ratio, i.e., the ratio of IoU on adversarial examples to that on clean point cloud over the whole validation set. The models that achieve a higher mAP ratio under adversarial attacks are considered more robust.
We assess the robustness of detectors against our presented adversarial attacks by measuring the mAP ratio. We report these values for both the KITTI and Waymo datasets.\\

\noindent\textbf{Perceptibility Measurement.}
We expect the adversarial examples to be imperceptible, i.e., they should be close to the corresponding raw point clouds. 
Therefore, we predominantly use the Chamfer Distance (CD) to measure the deviation of adversarial points $\mathbf{X}_a$ to the raw point cloud $\mathbf{X}$ (as described in Eq.~\eqref{eq:calculate_gradient}).\\

\color{black}
\noindent\textbf{Details of Defense.} 
The details of general defense used are as follows:
\begin{itemize}
	\item[$\bullet$] \textit{Scaling:} Each point cloud is uniformly scaled by a factor of 1.025.
	\item[$\bullet$] \textit{Rotation:} Point clouds are rotated 22.5 degrees clockwise along the $z$-axis.
	\item[$\bullet$] \textit{Gaussian Noise Addition:} A Gaussian noise with a standard deviation of 0.01 is added to the coordinates of each point within the cloud.
	\item[$\bullet$] \textit{Coordinate Quantization:} The coordinates of each point in the point cloud are quantized to the nearest centimeter.
	\item[$\bullet$] \textit{Data Augmentation:} We employ a stochastic data augmentation pipeline following the methodology proposed in PA-AUG~\citep{choi2021part}. This pipeline includes five operations: internal points dropout, cutmix, cutmixup, sparse sampling, and random noise generation.
	\item[$\bullet$] \textit{Random Subsampling:} 95\% of points are randomly sampled from the original point clouds.
	\item[$\bullet$] \textit{KNN-based Denoising:} The point clouds are partitioned into 20 clusters, and a 3-sigma rule is applied within each cluster to identify and remove outliers.
	\item[$\bullet$] \textit{Adversarial Training:} For point perturbation attacks, adversarial examples generated using PGD-based methods with a perturbation limit $\epsilon$ of 5 cm are used during training. For point detachment attacks, the training set includes adversarial examples where 2\% of points are detached. For point attachment attacks, adversarial examples are used that incorporate 164 additional points with a perturbation limit $\epsilon$ of 5 cm. From the perspective of training efficiency, we set the number of iterations to 1 for the iterative attacks to generate adversarial examples.
	\item[$\bullet$] \textit{BAFT:} We set the focusing parameter $\gamma=2.0$ and $\zeta=$0.5 to balance the loss on clean and adversarial examples.
\end{itemize}

\noindent\textbf{Details of Corruption Data.} We evaluate the corruption robustness of detectors by applying five common corruptions to the KITTI dataset in Section~\ref{sec:evaluation_corruption}. The implementation details of corruption are as follows:
\begin{itemize}
	\item[$\bullet$] \textit{Beam Missing:} The beam may not launch successfully when dust and insects form agglomerates in front of the LiDAR surface. Following \cite{kong2023robo3d}, we mimic the loss of impulses by randomly sampling 32 beams and dropping points on these beams.
	\item[$\bullet$] \textit{Snow:} We follow ~\cite{hahner2022lidar} to simulate the snow corruption. We set the value of the snowfall rate parameter to 0.5 to construct snow corruption.
	\item[$\bullet$] \textit{Drop Local:} 
	Because LiDAR sensors emit pulses in the near-infrared spectrum, it is difficult to accurately detect objects with dark colors or specific materials. This limitation can lead to an incomplete scan and significant omissions in the detected points. To simulate this corruption, we randomly sample 50\% points of each foreground object and remove them from the original point clouds.
	\item[$\bullet$] \textit{Jitter:} The LiDAR sensors inevitably suffer from the range error and the shaking of the car itself. Hence, we add noise subject to Gaussian distribution with a standard deviation of 0.06 to the coordinates of each point.
	\item[$\bullet$] \textit{Crosstalk:} LiDAR sensors operating in similar frequency ranges are prone to interfere with each other, causing discrepancies in point cloud data. Following \cite{kong2023robo3d}, we randomly select a subset comprising 0.8\% of the original point cloud. To these selected points, we introduce a substantial jittering noise. This noise is generated by translating the coordinates of the selected point based on a Gaussian distribution with a standard deviation of 3. Then we add the translated points to the original point clouds.
\end{itemize}
\color{black}

\setlength{\tabcolsep}{6pt}
\begin{table*}[htp]
	\centering
	\caption{Ablation of re-voxelization of perturbed point cloud and flipping the direction of perturbation imposed on points for the SECOND detector. \revise{We report the mAP ratio (\%) for comparison.}}
	\label{table:attack_towards_voxel_based_ablation}
	\begin{tabular}{c|cc|cccccc} 
		\toprule
		\multirow{2}{*}{Attack} & \multirow{2}{*}{Re-voxelization} & \multirow{2}{*}{Flipping Perturbation} & \multicolumn{6}{c}{$\epsilon$ (cm)}                        \\
		&  &     & 0.5 & 1   & 3   & 5   & 7   & 10    \\ 
		\hline
		\multirow{3}{*}{FGM}    &    &       & 99.98 & 99.54  & 99.39  & 99.19  & 98.62  & 98.44  \\
		& \checkmark   &       & 99.44 & 100.50 & 102.01 & 103.09 & 99.82  & 91.58  \\
		& \checkmark   & \checkmark      & 99.09 & 98.07  & 96.78  & 94.15  & 91.35  & 83.03  \\ 
		\hline
		\multirow{3}{*}{MI-FGM} &     &    & 99.98 & 99.54  & 99.39  & 99.21  & 98.63  & 98.40  \\
		& \checkmark    &     & 99.44 & 100.53 & 102.27 & 102.48 & 100.73 & 92.00  \\
		& \checkmark    & \checkmark    & 98.96 & 98.00  & 96.75  & 94.25  & 91.81  & 82.40  \\ 
		\hline
		\multirow{3}{*}{PGD}    &       &      & 99.98 & 99.60  & 99.47  & 99.35  & 99.20  & 98.63  \\
		& \checkmark            &           & 99.63 & 101.26 & 99.71  & 101.78 & 103.19 & 93.77  \\
		& \checkmark           & \checkmark         & 99.58 & 97.67  & 95.12  & 91.95  & 85.74  & 70.80  \\
		\bottomrule
	\end{tabular}
\end{table*}

\section{Evaluation of Adversarial Attacks}\label{sec:evaluation_of_adversarial_attacks}
In this section, we delve into a detailed assessment of various attack types and probe the resilience of different detectors to adversarial attacks.

\subsection{Probing Attacks on Voxel-based Detectors}\label{sec:attacking_voxel_based_detectors}
In our discussion from Section~\ref{sec:adversarial_point_perturbation}, we identified two core limitations of the point perturbation attack when it is performed on voxel-based detectors. In light of this, we undertook experiments to discern the efficacy of the point perturbation attack on these detectors. We also investigated the impact of re-voxelization operation and voxel size variability on the performance of attacks. From the quantitative results listed in Table~\ref{table:attack_towards_voxel_based_ablation}, we have the following observations:
\begin{enumerate}[1)]
    \item The point perturbation attack can hardly mislead the voxel-based detectors if the perturbed points are not re-voxelized. Without re-voxelization, the mAP ratios of SECOND are as high as 98.44\%, 98.40\%, and 98.63\% under FGM, MI-FGM, and PGD-based point perturbation, respectively. The results support our claim in Section~\ref{sec:adversarial_point_perturbation} that merely modifying the value of the voxel representation, without simultaneously adjusting its location, culminates in inferior attack outcomes.
   \item After applying re-voxelization of the perturbed point cloud, the performance of detectors even increases slightly. For example, the mAP ratio of SECOND is 103.19\% on the KITTI dataset for the PGD-based perturbation attack with $\epsilon$=7cm. A potential interpretation of these results is that the re-voxelization operation disrupts the unnatural patterns introduced by adversarial attacks and yields stronger robustness.
   \item Since the common attacks with the re-voxelization operation abnormally improve model performance, empirically, we further try to change the direction of perturbation, i.e., the sign of $\bigtriangledown_{\mathbf{X}_a^t} J(\mathbf{X}_a^t, \mathcal{G})$ in Eq.~\eqref{eq:adversarial_attack_iteration}, which is performed on the coordinates of points before re-voxelization. We can observe that attacks with flipped perturbation can effectively decrease the performance of detectors for all FGM, MI-FGM, and PGD-based point perturbation attacks with various $\epsilon$ values.
\end{enumerate}

\begin{table}[t]
	\centering
	\caption{Effects of perturbation direction adjustment during the point perturbation attack ($\epsilon$=10cm) on the SECOND detector across different voxel representation sizes. \revise{We report the mAP ratio (\%) for comparison.}}
	\label{table:flip_perturbation_for_different_voxel_sizes}
	\scalebox{0.85}{
		\begin{tabular}{c|c|ccc} 
			\toprule
			Voxel Size                            & Flipped Perturbation & FGM    & MI-FGM     & PGD     \\ 
			\midrule
			\multirow{2}{*}{{[}0.05, 0.05, 0.1]}  &               & 91.58  & 92.00  & 93.77   \\
			& \checkmark              & 83.03  & 82.40  & 70.80   \\
			\midrule
			\multirow{2}{*}{{[}0.10, 0.10, 0.2]}  &               & 105.73 & 104.30 & 101.10  \\
			& \checkmark              & 83.79  & 83.46  & 74.23   \\
			\midrule
			\multirow{2}{*}{{[}0.16, 0.16, 0.4]}  &               & 94.32  & 94.29  & 93.61   \\
			& \checkmark              & 96.84  & 94.04  & 97.37  \\
			\midrule
			\multirow{2}{*}{{[}0.20, 0.20, 0.4]~} &               & 88.64  & 91.65  & 89.55   \\
			& \checkmark              & 103.23 & 104.16 & 101.78   \\
			\bottomrule
		\end{tabular}
	}
\end{table}

\begin{figure*}[b]
	\centering
		\includegraphics[width=\textwidth]{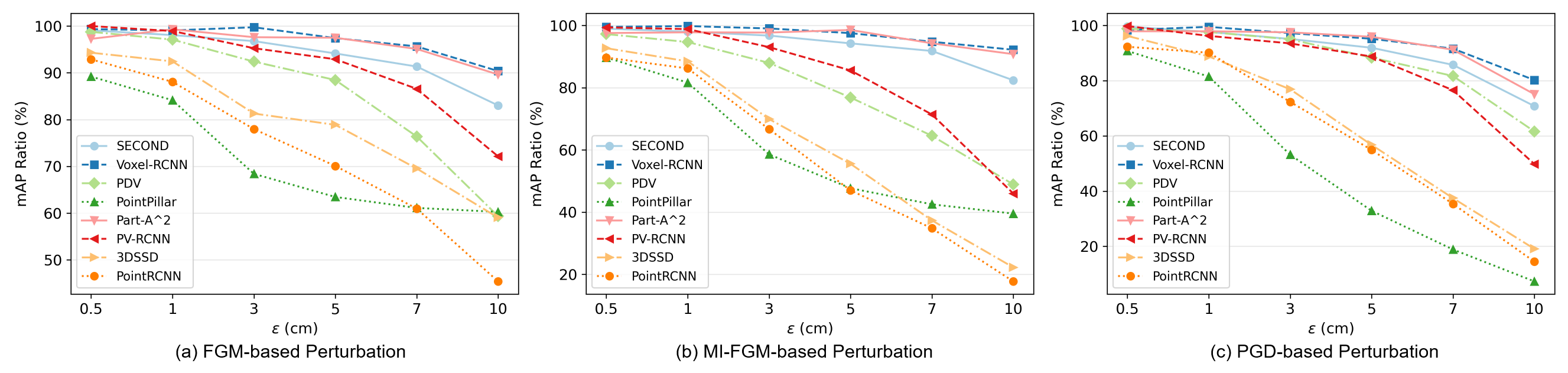} 
		\caption{\revise{Adversarial robustness of detectors against point perturbation attacks with different $\epsilon$ values on KITTI dataset.}}
		\label{fig:kitti_perturbation_attack}
\end{figure*}

\begin{figure*}[htp]
	\centering
		\includegraphics[width=\textwidth]{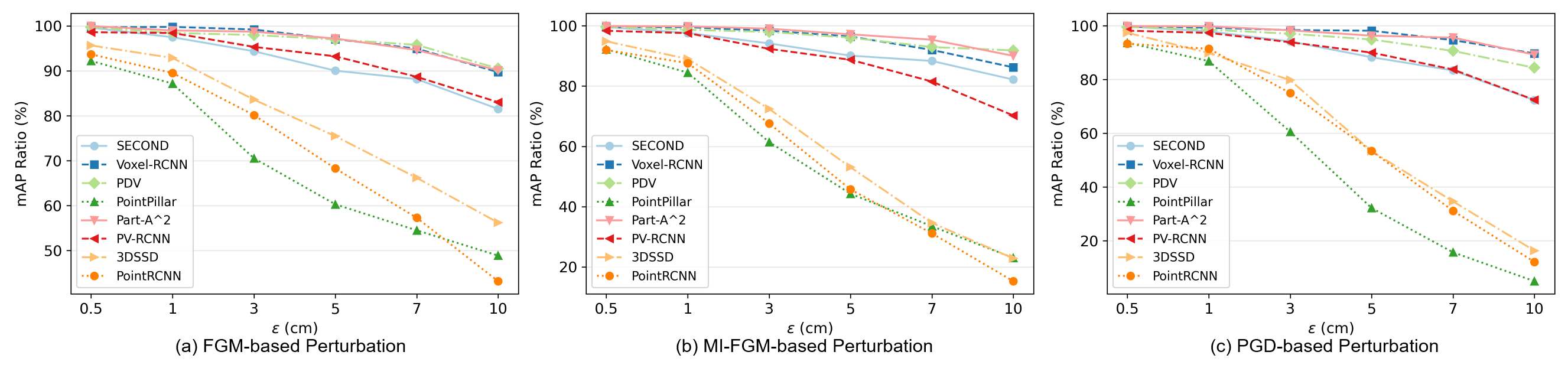} 
		\caption{\revise{Adversarial robustness of detectors against point perturbation attacks with different $\epsilon$ values on Waymo dataset.}}
		\label{fig:waymo_perturbation_attack}
\end{figure*}

Further, our research encompassed a deep dive into the influence of input voxel size on the robustness of voxel-based detectors. From Table~\ref{table:flip_perturbation_for_different_voxel_sizes}, we were able to deduce:

\begin{enumerate}[1)]
	\item The effectiveness of point perturbation attacks, when coupled with re-voxelization but devoid of flipping perturbations, increases for detectors that utilize larger voxel-based inputs. This observation lends credence to the hypothesis that re-voxelization dampens the patterns created by adversarial attacks. Consequently, the attack becomes more potent when the voxel size is enlarged, reducing the likelihood of perturbed points straying outside their original voxel boundaries.
	\item Conversely, when the voxel size is increased, making the point perturbation attack less sensitive to re-voxelization, altering the direction of perturbation as indicated in Eq.\eqref{eq:adversarial_attack_iteration} results in an augmented target value in Eq.\eqref{eq:gradient_optimize}. This, in turn, enhances the detection performance on adversarial examples.
\end{enumerate}

We also observe similar results on other voxel-based detectors including Voxel-RCNN, Part-A\^{}2. 
Given our commitment to maintaining the default configurations of detectors, we employed both re-voxelization and flipping perturbation tactics in subsequent sections, aiming for superior attack efficacy.

\subsection{Evaluation of Adversarial Point Perturbation Attack}\label{sec:evaluation_perturbation_attack}
Figure~\ref{fig:kitti_perturbation_attack} shows the robustness of several state-of-the-art detectors on the KITTI benchmark. From the reported results, we have the following observations:
\begin{enumerate}[1)]
\item In general, the most widely used gradient-based attack adopted for the perturbation of point clouds can successfully disrupt the 3D detectors. For example, the PGD-based point perturbation attack successfully reduces the mAP of the PointPillar from 67.22\% to 4.89\% (i.e., the mAP ratio is 72.82\%) under the PGD-based perturbation attack with $\epsilon$ of 10cm.

\item Most of the detectors achieve a similar mAP ratio under the setting of smaller $\epsilon$, while detectors show different levels of robustness against the point perturbation attack when the magnitude of the perturbation is larger.

\item As $\epsilon$ increases, the performance degradation of detectors becomes severe against three types of gradient-guided adversarial attacks. Iterative attacks (MI-FGM and PGD) are more effective than the single-step attack (FGM). On the most effective PGD-based attack (Figure~\ref{fig:kitti_perturbation_attack}(c)), the robustness of models varies significantly. Among the detectors, PointPillar, 3DSSD, and PointRCNN prove to be more vulnerable to adversarial examples.

\end{enumerate}

Besides, we also test the robustness of detectors on the Waymo validation set as shown in Figure~\ref{fig:waymo_perturbation_attack}. The detectors show similar relative rankings of robustness, which proves that the robustness is highly correlated with the structure and properties of the detector. We discuss the robustness of different networks in detail in Section~\ref{sec:robustness_of_networks}.

\begin{figure}[t]
	\centering
	\includegraphics[width=0.49\textwidth]{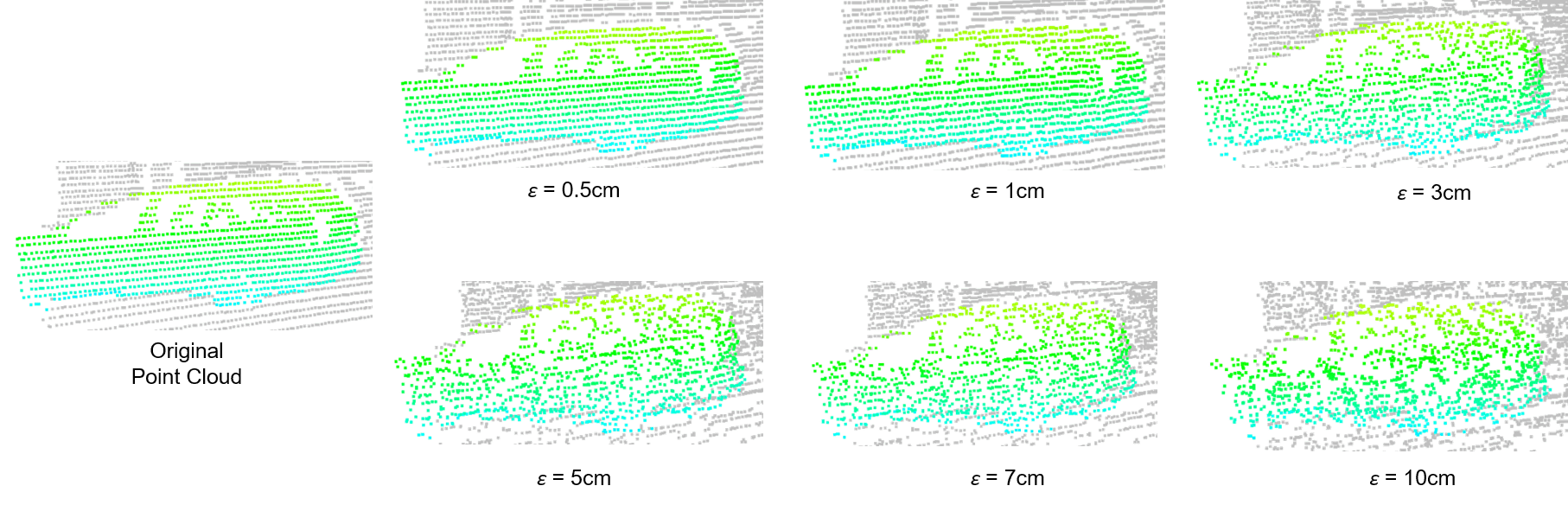} 
	\caption{
		Illustration of point clouds that belong to a typical car under point perturbation attacks with different $\epsilon$ values. It can be observed that the shape of the car is still preserved under adversarial attack.
	}
	\label{fig:deformation_versus_epsilon}
\end{figure}

%
Figure~\ref{fig:deformation_versus_epsilon} illustrates the deformation experienced by a point cloud when subjected to point perturbation attacks with different values of $\epsilon$. Although a higher $\epsilon$ value enhances the effectiveness of the attack, it also causes the point cloud to deviate significantly from its original shape. To ensure the practicality of our experiments and prevent unrealistic deformations, we refrain from using an $\epsilon$ value greater than 10cm for point perturbation attacks. Besides, we provide more qualitative results in Appendix \textcolor{red}{B}.


\subsection{Evaluation of Adversarial Point Detachment Attack}
\begin{figure*}[htbp]
	\centering
		\includegraphics[width=\textwidth]{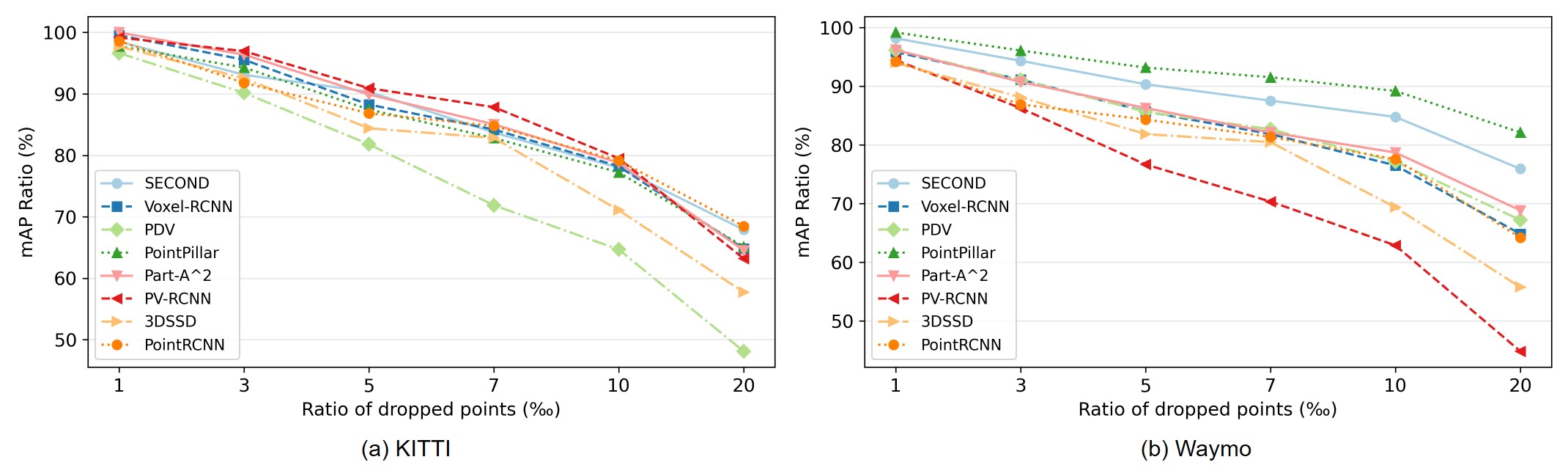} 
		\caption{
			\revise{Adversarial robustness of detectors against point detachment attacks with different ratios of detached points on KITTI and Waymo datasets.}
		}
		\label{fig:detach_attack}
\end{figure*}

Figure~\ref{fig:detach_attack} compares the performance of detectors with respect to different ratios of dropped points when the point detachment attack is applied. Intuitively, the performance decreases when the ratio of dropped points increases. For instance, the performance of PDV dropped by 51.87\% (i.e., 48.13\% mAP ratio) under the point detachment attack with only 2\% points dropped. The low performance of the detectors under adversarial point detachment attack is understandable since it is hard to predict 3D bounding boxes with seven degrees of freedom based on fewer points.
For the adversarial detachment attack, the margin between the most and least robust network is smaller compared to that of the point perturbation attack. The models including SECOND, 3DSSD, and PointRCNN are clearly less vulnerable to the point detachment attack compared with the point perturbation attack. We further evaluate nine defense strategies against point detachment attacks in Section~\ref{sec:eval_defense}.

\begin{figure*}[htp]
	\centering
		\includegraphics[width=\textwidth]{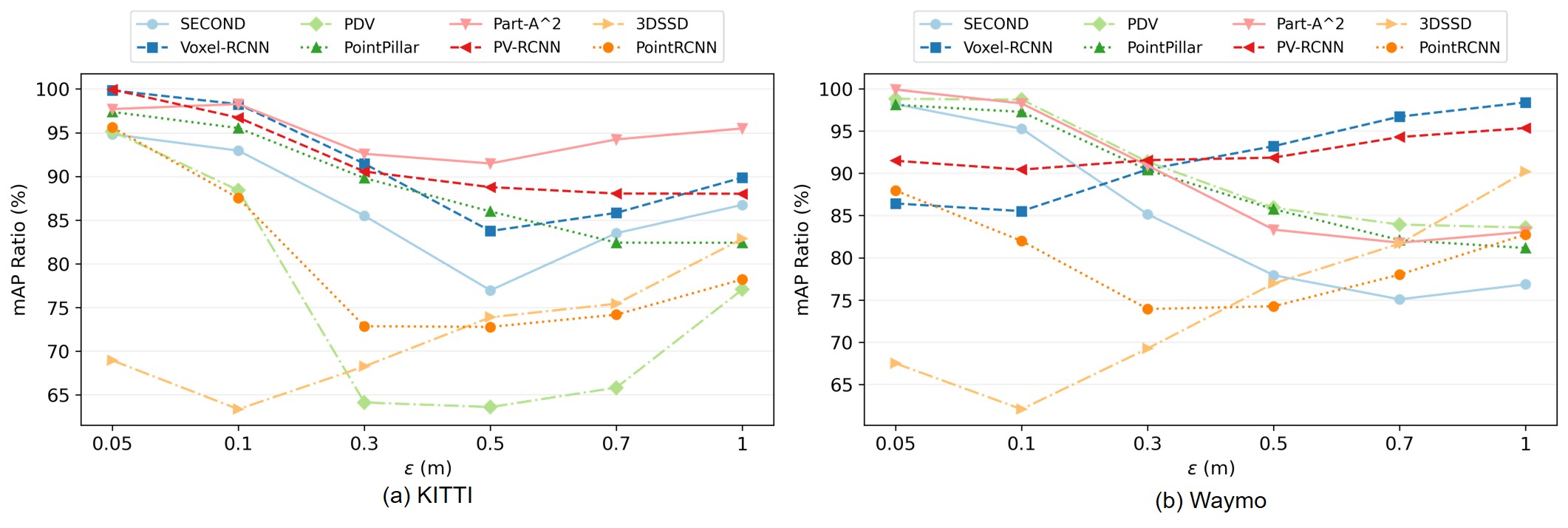} 
		\caption{
			\revise{Adversarial robustness of detectors against point attachment attacks with different values of $\epsilon$ for attached points. The number of attached points is set to 164 and 500 on the KITTI and Waymo datasets, respectively.}
		}
		\label{fig:attach_attack}
\end{figure*}

\subsection{Evaluation of Adversarial Point Attachment Attack}
Figure~\ref{fig:attach_attack} shows the performance of detectors against the point attachment attacks with respect to different $\epsilon$ values. From this analysis, we gather the following insights:
\begin{enumerate}[1)]
\item It is interesting to see that the performance of detectors does not necessarily deteriorate with an increase in $\epsilon$.
This intriguing phenomenon suggests that when $\epsilon$ is larger, the newly attached points are displaced further away. As a consequence, these additions might be less impactful on crucial points, resulting in reduced interference.
\item Compared with the performance of the point perturbation attack, the results reported in Figure~\ref{fig:attach_attack} imply that attacking vulnerable regions by adding new points is less malignant than directly applying perturbation to the original point cloud.
\item The PDV detector clearly exhibits susceptibility to both point attachment and detachment attacks on the KITTI dataset. This vulnerability in PDV could be attributed to its unique design. Specifically, PDV utilizes point density features and incorporates a density confidence prediction module. This module refines confidence predictions for proposals by considering the \textit{number} of LiDAR points contained within each proposal as additional features~\citep{hu2022point}.
\end{enumerate}


\setlength{\tabcolsep}{2pt}
\begin{table}[t]
	\centering
	\caption{Performance of three attacks towards different detectors and the corresponding perceptibility of adversarial examples on the KITTI dataset. The mAP ratio (\%) is reported.}
	\label{table:performance_versus_perceptibility}
	\scalebox{0.9}{
		\begin{threeparttable}
			\begin{tabular}{l|cc|cc|cc}
				\toprule
				\multirow{2}{*}{Detectors} & \multicolumn{2}{c|}{Perturbation} & \multicolumn{2}{c|}{Detachment} & \multicolumn{2}{c}{Attachment}  \\
				& CD     & mAP Ratio                & CD     & mAP Ratio              & CD     & mAP Ratio              \\ 
				\midrule
				SECOND                     & 0.0125 & 70.80                    & 0.0260 & 67.92                  & 0.0004 & 85.50                  \\
				Voxel-RCNN                 & 0.0124 & 80.24                    & 0.0107 & 64.84                  & 0.0004 & 91.45                  \\
				PDV                        & 0.0125 & 61.61                    & 0.0267 & 48.13                  & 0.0005 & 64.14                  \\
				PointPillar                & 0.0125 & 7.28                     & 0.1116 & 65.14                  & 0.0001 & 89.80                  \\
				Part-A\^{}2                     & 0.0124 & 75.11                    & 0.0074 & 64.54                  & 0.0004 & 92.58                  \\
				PV-RCNN                    & 0.0125 & 49.80                    & 0.0183 & 63.21                  & 0.0004 & 90.56                  \\
				3DSSD                      & 0.0129 & 19.07                    & 0.0222 & 57.75                  & 0.0005 & 68.27                  \\
				PointRCNN                  & 0.0142 & 14.44                    & 0.0100 & 68.49                  & 0.0005 & 72.86                  \\ 
				\midrule
				Average                    & 0.0127 & 47.29                    & 0.0291 & 62.50                  & 0.0004 & 81.89                  \\
				\bottomrule
			\end{tabular}
			\begin{tablenotes}
				\item Note: $\epsilon$ is set to 10cm and 30cm for point perturbation and attachment attack, respectively. The ratio of dropped points is set to 2\% for point detachment attacks.
			\end{tablenotes}
		\end{threeparttable}
	}
\end{table}
\setlength{\tabcolsep}{6pt}

\subsection{Attack Performance Versus Perceptibility}
Table~\ref{table:performance_versus_perceptibility} presents the performance of adversarial attacks on various detectors, as well as the perceptibility of the corresponding adversarial examples. It can be observed the point perturbation attack achieves the best attack performance while the perceptibility of its corresponding adversarial examples is lower than the point detachment attack.
Although the adversarial examples generated by the point attachment attack are least perceptible, the attack is far less effective than point perturbation and detachment attacks.    

With the constraint of $\epsilon$ in Eq.~\eqref{eq:perturbation_clip} to limit the magnitude of perturbation, the CD of adversarial examples is similar among different detectors under point perturbation attack.

Additionally, we find that the effectiveness of attacks is not directly proportional to the CD between the adversarial examples and the raw point cloud. For example, for PointPillar, the performance (7.28\% mAP ratio) under the point perturbation attack is better than that of the point detachment attack (65.14\%), though the CD metric of the former is smaller.

\color{black}
\subsection{Ablation Study on PointPillar}\label{sec:ablation_pointpillar}
As mentioned in Section~\ref{sec:evaluation_perturbation_attack}, PointPillar is more vulnerable to attacks than traditional voxel-based detectors. 
First, while traditional voxel-based methods typically divide the 3D space into small cubic voxels, PointPillars utilizes a larger voxel size, segmenting the space into vertical columns called pillars. Second, PointPillars avoids the computational overhead of intricate 3D convolutions. The third distinction lies in the encoding of features within each voxel or pillar. Specifically, for PointPillars, the feature set of each point within a pillar encompasses not just the absolute coordinates and intensity, but also the relative distance to the centroid of the pillar and the centroid of points within that pillar.

In this section, we explore how the aforementioned three factors influence the robustness of PointPillar against point perturbation attacks. First, we conduct ablative experiments on the detectors with different pillar sizes. As shown in Table~\ref{table:ablation_pointpillar_pillar_size}, we can observe that changing the size of the pillar in the dimensions of $x$ and $y$ does not improve the robustness of PointPillar. Therefore, the voxel size is not the key factor of the robustness of PointPillar.

\setlength{\tabcolsep}{15pt}
\begin{table}[htp]
	\centering
	\caption{
		\revise{Effect of the pillar size on the robustness of PointPillar under PGD-based point perturbation attack on the KITTI dataset. The mAP ratio (\%) is reported.}	
	}
	\label{table:ablation_pointpillar_pillar_size}
	\begin{tabular}{c|cc} 
		\toprule
		\multirow{2}{*}{Pillar Size} & \multicolumn{2}{c}{$\epsilon$ (cm)}  \\
		& 5     & 10              \\ 
		\hline
		{[}0.08, 0.08, 4]                        & 32.97 & 6.79            \\
		{[}0.10, 0.10, 4]                        & 27.08 & 4.37            \\
		{[}0.12, 0.12, 4]                        & 33.80 & 7.53            \\
		{[}0.14, 0.14, 4]                        & 28.62 & 1.84            \\
		{[}0.16, 0.16, 4]                        & 32.87 & 7.28            \\
		{[}0.18, 0.18, 4]                        & 13.68 & 1.76            \\
		{[}0.20, 0.20, 4]                        & 27.65 & 4.96            \\
		\bottomrule
	\end{tabular}
\end{table}
\setlength{\tabcolsep}{6pt}

\setlength{\tabcolsep}{8pt}
\begin{table}[htp]
	\centering
	\caption{
		\revise{Effect of the feature encoding of the pillar on the robustness of PointPillar under PGD-based point perturbation attack on the KITTI dataset. The mAP ratio (\%) is reported. ``Abs. \& Intensity" denotes the absolute coordinates and intensity of points in the pillar. The $D_{\text{cluster}}$ and $D_{\text{center}}$ are the distance of points to the centroid of points within the pillar and the center of pillar, respectively.}
	}
	\label{table:ablation_pointpillar_encoding}
	\begin{tabular}{ccc|cc} 
		\toprule
		\multirow{2}{*}{Abs. \& Intensity} & \multirow{2}{*}{$D_{\text{cluster}}$} & \multirow{2}{*}{$D_{\text{center}}$} & \multicolumn{2}{c}{$\epsilon$ (cm)}  \\
		&        &     & 5        & 10           \\ 
		\hline
		\checkmark          &        &        & 83.77 & 64.25  \\
		\checkmark       & \checkmark       &          & 49.65 & 25.24  \\
		\checkmark        &       & \checkmark       & 36.50 & 11.59  \\
		\checkmark         & \checkmark   & \checkmark        & 32.86 & 7.28  \\
		\bottomrule
	\end{tabular}
\end{table}
\setlength{\tabcolsep}{6pt}
Next, we explore the effect of feature encoding of pillar. Table~\ref{table:ablation_pointpillar_encoding} shows that pillar encoding significantly impacts the robustness of PointPillar. Incorporating local features, i.e., the distance of points to the centroid of points and the pillar center, deteriorates the robustness of PointPillar. When only the absolute coordinates and intensity of points are used, the PointPillar can achieve 64.25\% mAP ratio under the PGD-based point perturbation attack with $\epsilon$ of 10cm, which is comparable to the robustness of traditional voxel-based 3D detectors.
Therefore, we can conclude the vulnerability of PointPillar results from the special encoding of the pillar, rather than the large size of pillar or the absence of 3D convolution layers.

\subsection{Comparison of Results on KITTI and Waymo}\label{sec:comparison_kitti_waymo}
An analysis of Fig.\ref{fig:kitti_perturbation_attack} and Fig.\ref{fig:waymo_perturbation_attack} reveals that detectors exhibit greater robustness on the Waymo dataset than on the KITTI dataset when subjected to point perturbation. For example, the mAP ratio of PDV during a PGD-based perturbation attack with an $\epsilon$ of 10cm stands at 61.61\% for KITTI and 84.37\% for Waymo. This trend is similarly observed with point detachment attacks, as seen in Fig.~\ref{fig:detach_attack}. As the model structure remains consistent across both datasets, the higher resolution of the Waymo dataset (as described in Section~\ref{sec:dataset}) 
contribute to this discrepancy. Hence, we infer that point cloud quality plays a crucial role in determining the robustness of 3D detectors, with models exhibiting enhanced robustness when trained on high-quality point clouds.

\color{black}

\subsection{The Robustness of Different Networks}\label{sec:robustness_of_networks}
\begin{figure}[t]
	\centering
	\includegraphics[width=0.5\textwidth]{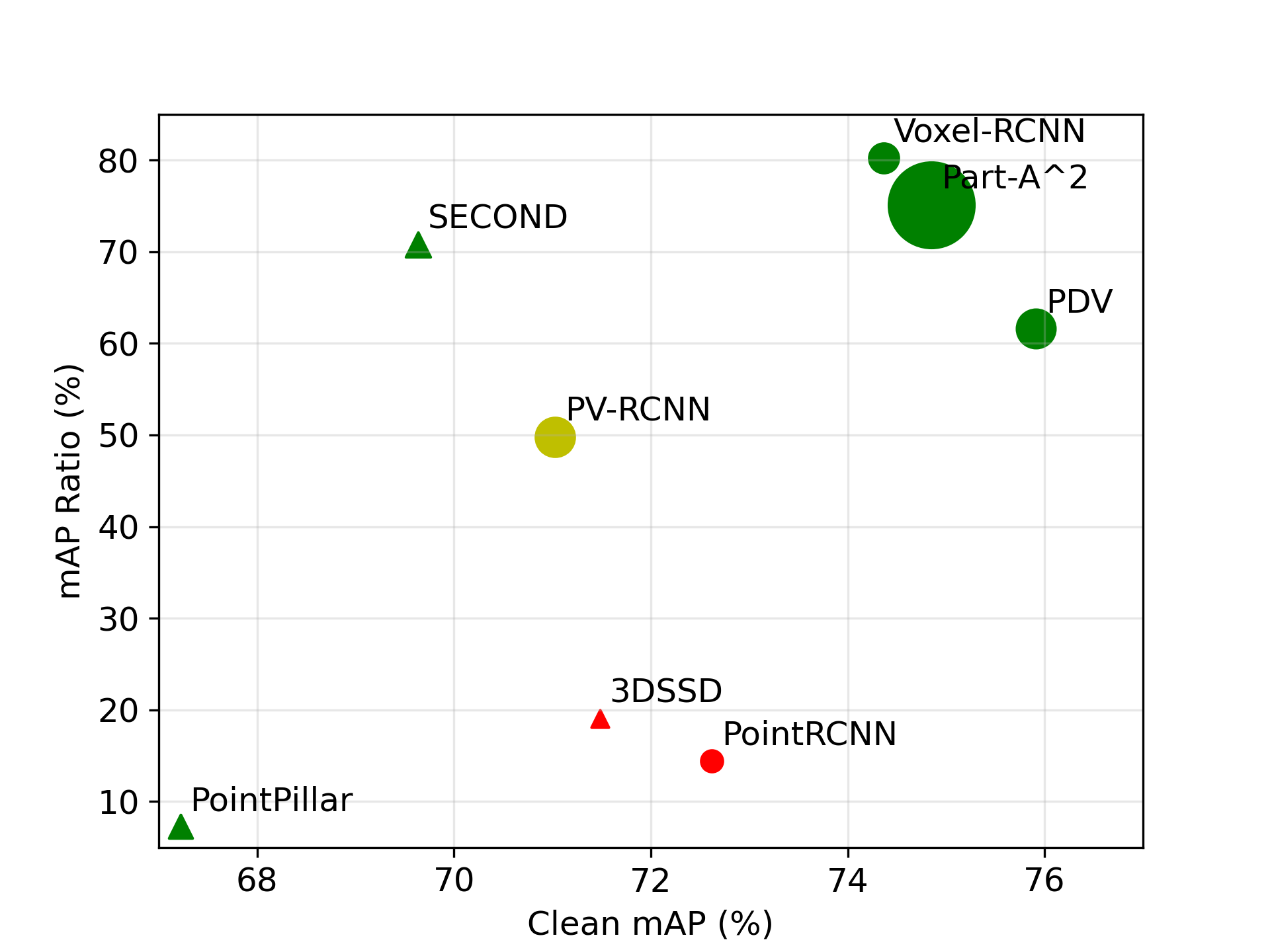} 
	\caption{
		The mAP ratio of detectors under the PGD-based point perturbation attack with $\epsilon$=10cm on the KITTI dataset. We use red, green, and yellow to denote the \textcolor{red}{voxel-based}, \textcolor{green}{point-based}, and {\color{Goldenrod}{point-voxel-based detectors}}, respectively. The triangle and circle denote the one-stage and two-stage detectors respectively. The larger the marker, the larger the network size.
	}
	\label{fig:ratio_versus_clean}
\end{figure}
The point perturbation-based attack has been demonstrated to be the most effective among the three types of attacks. 
We further illustrate the correlation between the mAP on clean data and robustness to the \textit{perturbation attack} in Figure~\ref{fig:ratio_versus_clean}. From this, we make the following observations about the effect of network structure on robustness:
\begin{enumerate}[1)]
    \item First, voxel-based methods generally achieve higher mAP ratios against the adversarial perturbation attack than the point-based detectors. Both point-based 3DSSD and PointRCNN show much lower mAP ratios than voxel-based detectors except for PointPillar. However, we cannot conclusively state that voxel-based detectors are more robust. As stated in Section~\ref{sec:adversarial_point_perturbation}, there are two limitations when applying point perturbation attack against voxel-based detectors, which make the adversarial examples generated from the voxel-based method less malignant. 

   \item  PDV achieves the highest performance on the clean point cloud input but it shows worse robustness compared with the simple SECOND detector. This indicates that detection accuracy on clean input does not equate to robustness; models with higher accuracy may actually be less resistant to adversarial attacks. These results contrast with the observed correlation between accuracy on clean data and robustness in image-based tasks~\citep{szegedy2013intriguing}.


   \item The lower mAP ratio achieved by PV-RCNN, compared to voxel-based two-stage detectors like Voxel-RCNN and Part-A\^{}2, is likely due to the point-voxel representation introduced by the multi-scale point-voxel integration strategy.

   \item Compared with feature representation and other properties, model size \textit{does not} significantly affect the robustness against adversarial attacks. For example, SECOND and PointRCNN with similar sizes of parameters (5.33M vs 4.04M) show largely different robustness. This phenomenon indicates that some observations made in the context of image classification, such as too large or too small models leading to poor robustness~\citep{kurakin2017adversarial}, cannot be directly transferred to the task of 3D object detection.

   \item The robustness is not significantly affected by whether the model uses a one-stage or two-stage structure when compared to feature representation.
\end{enumerate}


\setlength{\tabcolsep}{6pt}
\begin{table*}[htp]
	\renewcommand\arraystretch{1.5}
	\centering
	\caption{Transferability of adversarial examples for \textbf{point perturbation attack} generated from source models and evaluated on target models. The gray diagonals denote white-box attacks, and off-diagonals show transfer (black box) attacks. The most effective black-box attack for each target detector is bold.}
	\label{table:perturbation_trans}
	\begin{tabular}{cc|cccccccc} 
		\toprule
		\multicolumn{2}{l|}{\multirow{2}{*}{mAP Ratio (\%)}} & \multicolumn{8}{c}{\underline{Source Model}}                                          \\
		\multicolumn{2}{l|}{}     & SECOND & Voxel-RCNN & PDV   & PointPillar & Part-A\^{}2 & PV-RCNN & 3DSSD & PointRCNN  \\
		\midrule
		\multirow{8}{*}{\rotatebox{90}{\underline{Target Model}}} & SECOND              & \cellcolor{light-gray}70.80  & 77.22      & 65.55 & 65.65       & 77.36  & 64.99   & 54.04 & \textbf{49.77}      \\
		& Voxel-RCNN          & 85.06  & \cellcolor{light-gray}80.24      & 71.64 & 71.20       & 86.90  & 72.78   & 67.26 & \textbf{58.47}      \\
		& PDV                 & 83.69  & 86.90      & \cellcolor{light-gray}61.61 & 74.05       & 86.35  & 75.50   & 66.80 & \textbf{59.69}      \\
		& PointPillar         & 92.64  & 93.17      & 84.00 & \cellcolor{light-gray}7.28        & 95.44  & 88.37   & \textbf{80.48} & 81.29      \\
		& Part-A\^{}2              & 80.65  & 80.40      & 63.32 & 63.62       & \cellcolor{light-gray}75.11  & 65.73   & 58.96 & \textbf{52.59}      \\
		& PV-RCNN             & 75.09  & 76.55      & 59.92 & 63.22       & 75.85  & \cellcolor{light-gray}47.96   & 58.29 & \textbf{50.65}      \\
		& 3DSSD               & 87.32  & 85.46      & 73.70 & 77.53       & 88.55  & 73.55   & \cellcolor{light-gray}19.07 & \textbf{55.52}     \\
		& PointRCNN           & 71.78  & 70.95      & 59.23 & 64.55       & 71.39  & 59.79   & \textbf{56.67} & \cellcolor{light-gray}14.44      \\
		\bottomrule
	\end{tabular}
\end{table*}

\setlength{\tabcolsep}{6pt}
\begin{table*}[htp]
	\renewcommand\arraystretch{1.5}
	\centering
	\caption{Transferability of adversarial examples for \textbf{point detachment attack} generated from source models and evaluated on target models. The gray diagonals denote white-box attacks, and off-diagonals show transfer (black box) attacks. The most effective black-box attack for each target detector is bold.}
	\label{table:detachment_trans}
	\begin{tabular}{cc|cccccccc} 
		\toprule
		\multicolumn{2}{l|}{\multirow{2}{*}{mAP Ratio (\%)}} & \multicolumn{8}{c}{\underline{Source Model}}                                          \\
		\multicolumn{2}{l|}{}    & SECOND & Voxel-RCNN & PDV   & PointPillar & Part-A\^{}2 & PV-RCNN & 3DSSD & PointRCNN  \\
		\midrule
		\multirow{8}{*}{\rotatebox{90}{\underline{Target Model}}} & SECOND     & \cellcolor{light-gray}67.92 & 63.99      & \textbf{47.90} & 62.06       & 64.63  & 59.91   & 55.40 & 68.71      \\
		& Voxel-RCNN & 68.39 & \cellcolor{light-gray}64.84      & \textbf{49.47} & 63.48       & 65.34  & 62.27   & 54.02 & 67.98      \\
		& PDV        & 65.49 & 64.01      & \cellcolor{light-gray}48.13 & 63.26       & 64.95  & 59.19   & \textbf{53.87} & 67.78      \\
		& PointPillar                      & 67.79 & 63.63      & \textbf{49.42} & \cellcolor{light-gray}65.14       & 63.93  & 57.04   & 54.77 & 64.73      \\
		& Part-A\^{}2     & 67.58 & 64.08      & \textbf{49.11} & 60.47       & \cellcolor{light-gray}64.54  & 61.25   & 53.60 & 70.12      \\
		& PV-RCNN    & 69.66 & 65.31      & \textbf{50.78} & 64.25       & 66.92  & \cellcolor{light-gray}63.21   & 55.43 & 71.26      \\
		& 3DSSD      & 70.19 & 66.34      & \textbf{53.14} & 66.75       & 68.80  & 59.69   & \cellcolor{light-gray}57.75 & 74.06      \\
		& PointRCNN  & 61.73 & 58.89      & \textbf{43.85} & 60.10       & 60.52  & 51.60   & 56.33 & \cellcolor{light-gray}68.49      \\
		\bottomrule
	\end{tabular}
\end{table*}

\setlength{\tabcolsep}{6pt}
\begin{table*}[htp]
	\renewcommand\arraystretch{1.5}
	\centering
	\caption{Transferability of adversarial examples for \textbf{point attachment attack} generated from source models and evaluated on target models. The gray diagonals denote white-box attacks, and off-diagonals show transfer (black box) attacks. The most effective black-box attack for each target detector is bold.}
	\label{table:attachment_trans}
	\begin{tabular}{cc|cccccccc} 
		\toprule
		\multicolumn{2}{l|}{\multirow{2}{*}{mAP Ratio (\%)}} & \multicolumn{8}{c}{\underline{Source Model}}                                          \\
		\multicolumn{2}{l|}{}    & SECOND & Voxel-RCNN & PDV   & PointPillar & Part-A\^{}2 & PV-RCNN & 3DSSD & PointRCNN  \\
		\midrule
		\multirow{8}{*}{\rotatebox{90}{\underline{Target Model}}} & SECOND      & \cellcolor{light-gray}85.50  & 88.87  & \textbf{71.30} & 94.24   & 88.77  & 92.04   & 86.96 & 90.21  \\
		& Voxel-RCNN  & 90.55  & \cellcolor{light-gray}91.45  & \textbf{72.51} & 96.31   & 92.65  & 93.17   & 88.68 & 89.90  \\
		& PDV & 89.67  & 90.77  & \cellcolor{light-gray}64.14 & 95.54   & 91.91  & 92.03   & \textbf{89.27} & 90.21  \\
		& PointPillar & 87.61  & 87.92  & \textbf{75.51} & \cellcolor{light-gray}89.80   & 91.31  & 92.79   & 83.21 & 84.83  \\
		& Part-A\^{}2  & 91.71  & 91.15  & \textbf{75.80} & 94.60   & \cellcolor{light-gray}92.58  & 93.17   & 92.07 & 92.09  \\
		& PV-RCNN & 91.67  & 93.76  & \textbf{71.51} & 96.93   & 92.86  & \cellcolor{light-gray}90.56   & 91.63 & 94.42  \\
		& 3DSSD   & 88.92  & 86.08  & \textbf{68.30} & 95.06   & 91.67  & 97.43   & \cellcolor{light-gray}68.27 & 96.95  \\
		& PointRCNN   & 78.25  & 77.49  & \textbf{61.79} & 88.43   & 79.56  & 79.73   & 87.91 & \cellcolor{light-gray}72.86  \\
		\bottomrule
	\end{tabular}
\end{table*}

\section{Transferability Evaluation of Attacks}\label{sec:transferability_evaluation}
In the preceding section, we assessed the effectiveness of the source attack, in which the adversarial examples are obtained with available models and known ground-truth bounding boxes. As direct access to the target detectors might not always be available, we also explore the potential vulnerability of detectors under black-box attacks. Except for the source attack assessed, we further study the effectiveness of attacks in three settings in this section:
\begin{itemize}
	\item[$\bullet$] \textit{Cross-model Attack}: We obtain the adversarial examples learned from another model to fool the target detectors.
	\item[$\bullet$] \textit{Cross-domain Attack}: The adversarial examples are generated based on detectors that are optimized on another domain, which are used to fool the detectors trained on the target domains.
	\item[$\bullet$] \textit{Cross-task Attack}: We use the adversarial examples generated on 3D semantic segmentation tasks to attack the detectors.
\end{itemize}
The source attack is categorized as a white-box attack, while the cross-model, cross-domain, and cross-task attacks belong to the black-box setting, since the structure and parameters of the targeted 3D detectors are not necessarily available in those attacks.

\subsection{Evaluation on Cross-model Attack}
Table~\ref{table:perturbation_trans} summarizes the transferability of PGD-based point perturbation attack for the eight kinds of detectors on the KITTI dataset, where $\epsilon$=10cm. From the results, we have the following observations:
\begin{enumerate}[1)]
	\item Adversarial examples exhibit some degree of transferability across various detectors. We can clearly observe the adversarial examples generated based on vulnerable point-based detectors, i.e. PointRCNN and 3DSSD, are also effective in attacking other detectors. Specifically, the detectors including SECOND, Voxel-RCNN, PDV, and Part-A\^{}2 achieve lower mAP ratios under the transferred attack than that under the source attack when the PointRCNN is used as the source model. 
	As analyzed in Section~\ref{sec:attacking_voxel_based_detectors}, re-voxelization for voxel-based detectors dampens the patterns created by adversarial attacks. 
	Despite our efforts to address this by altering the perturbation direction (as discussed in Section~\ref{sec:attacking_voxel_based_detectors}), these manipulations remain less effective than transferred attacks from point-based detectors.

	\item 
	Notably, even though PointPillar is vulnerable to point perturbation attacks, it does not produce adversarial examples that significantly affect other detectors. Similarly, examples from other detectors demonstrate limited attack efficacy against PointPillar. This peculiarity can be attributed to the unique design of PointPillar, incorporating a distinct voxel feature encoding (see Section~\ref{sec:ablation_pointpillar}) and the Pillar feature network, which differentiates it from other detection mechanisms.
	
\end{enumerate}

\begin{table*}[b]
	\centering
	\caption{
		Performance of \textbf{cross-task} point perturbation attack that uses perturbation generated from 3D semantic segmentation model to attack 3D object detectors. The mAP ratio (\%) is reported.
	}
	\label{table:cross_task_trans}
	\begin{tabular}{c|c|cccccccc|c} 
		\toprule
		\multicolumn{2}{c|}{Attack}      & SECOND & Voxel-RCNN & PDV   & PointPillar & Part-A\^{}2 & PV-RCNN & 3DSSD & PointRCNN & Average  \\
		\midrule
		\multirow{3}{*}{\makecell[c]{Source \\ Attack}}     & FGM    & 83.03  & 90.38      & 59.35 & 61.09       & 89.64   & 72.14   & 59.05 & 45.40     & 70.01    \\
		& MI-FGM & 82.40  & 92.20      & 48.93 & 39.55       & 90.79   & 46.02   & 22.27 & 17.76     & 54.99    \\
		& PGD    & 70.80  & 80.24      & 61.61 & 7.28        & 75.11   & 49.80   & 19.07 & 14.44     & 47.29    \\
		\midrule
		\multirow{3}{*}{\makecell[c]{Cross-task \\ Attack}} & FGM    & 59.35  & 70.77      & 68.53 & 86.18       & 60.62   & 54.07   & 69.39 & 66.70     & 66.95    \\
		& MI-FGM & 41.92  & 54.63      & 54.97 & 76.56       & 44.06   & 37.68   & 52.69 & 51.68     & 51.77    \\
		& PGD    & 48.09  & 59.95      & 61.72 & 80.23       & 47.43   & 43.20   & 62.15 & 56.88     & 57.46    \\
		\bottomrule
	\end{tabular}
\end{table*}

Table~\ref{table:detachment_trans} summarizes the transferability of the \textit{point detachment attack} generated from different detectors on the KITTI dataset, where the ratio of detached points is set to 2\%. Adversarial examples, originating from PDV and 3DSSD (both vulnerable to the source attack), demonstrate enhanced attack performance against other detectors. In general, detectors show similar robustness against the transferred point detachment attack generated from another detector.

Besides, Table~\ref{table:attachment_trans} summarizes the transferability of \textit{point attachment attack} generated from different detectors on the KITTI dataset, where the max displacement of attached points $\epsilon$ is set to $30$cm. We can observe that the transferred point attachment attack generated from PDV, which shows vulnerability to the source attack, is most effective on other detectors. The heightened vulnerability of PDV, in comparison to other voxel-based detectors, could stem from its use of Density Confidence Prediction~\citep{hu2022point}. This technique refines confidence predictions of proposals by incorporating the number of LiDAR points within them as additional features, potentially making the pipeline more sensitive to point detachments or attachments in critical areas.


\subsection{Evaluation on Cross-task Attack}
In this part, we investigate whether the adversarial examples generated on 3D semantic segmentation models can transfer to 3D detectors. Specifically, we adopt a PointNet++ structure for the 3D semantic segmentation and obtain adversarial examples using the point perturbation attack described in Section~\ref{sec:adversarial_attacks} and adapted with the loss of 3D semantic segmentation. We conduct experiments under the point perturbation attack instead of point detachment and attachment, considering its appropriate balance between attack ability and perceptibility reported in Table~\ref{table:performance_versus_perceptibility}.

We summarize the results of cross-task transfer attacks in Table~\ref{table:cross_task_trans}. Adversarial examples are generated using FGM, MI-FGM, and PGD rule ($\epsilon$=10cm) on semantic segmentation models and evaluated on state-of-the-art detectors. Generally, the cross-task transfer point perturbation attack is significantly more effective than the direct source attack reported in Figure~\ref{fig:kitti_perturbation_attack} on voxel-based detectors. Specifically, the mAP ratios under source point perturbation attack (PGD-based) of SECOND, Voxel-RCNN, Part-A\^{}2 are 70.80\%, 80.24\%, and 75.11\%, respectively, while that under transferred attack from 3D semantic segmentation are 48.09\%, 59.95\%, and 47.43\%. 
This observation suggests the weakness of source attacks for voxel-based detectors again. 
We cannot assume that voxel-based detectors are inherently robust merely because it has higher mAP ratios under source attack, considering the inherent limitation of source attack as described in Section~\ref{sec:attacking_voxel_based_detectors}. Even so, considering the performance of detectors under source attack, cross-model attack, and cross-task attack, voxel-based detectors are still generally more robust than point-based detectors. 
By the way, the iterative attacks are still more effective than the single-step attacks in the cross-task attack.

\begin{table*}[t]
	\centering
	\caption{
		Performance of \textbf{cross-domain} point perturbation attack that uses adversarial examples generated from models trained on Waymo to attack target model on KITTI dataset. The cross-domain attack belongs to the black-box attack since it is unnecessary to access the parameters of the target model. The mAP ratio (\%) is reported.
	}
	\label{table:cross_domain_trans}
	\begin{tabular}{c|c|cccccccc|c} 
		\toprule
		\multicolumn{2}{c|}{Attack}     & SECOND & Voxel-RCNN & PDV   & PointPillar & Part-A\^{}2 & PV-RCNN & 3DSSD & PointRCNN & Average  \\
		\midrule
		\multirow{3}{*}{\makecell[c]{Source \\ Attack}}     & FGM    & 83.03  & 90.38      & 59.35 & 61.09       & 89.64   & 72.14   & 59.05 & 45.40     & 70.01    \\
		& MI-FGM & 82.40  & 92.20      & 48.93 & 39.55       & 90.79   & 46.02   & 22.27 & 17.76     & 54.99    \\
		& PGD    & 70.80  & 80.24      & 61.61 & 7.28        & 75.11   & 49.80   & 19.07 & 14.44     & 47.29    \\
		\midrule
		\multirow{3}{*}{\makecell[c]{Cross-domain \\ Attack}} & FGM    & 97.69  & 71.53      & 75.60 & 91.08       & 71.22   & 85.91   & 75.18 & 67.44     & 79.46    \\
		& MI-FGM & 98.20  & 47.80      & 75.04 & 93.77       & 59.75   & 85.57   & 44.17 & 41.68     & 68.24    \\
		& PGD    & 84.03  & 54.62      & 66.34 & 90.69       & 64.10   & 78.12   & 37.07 & 40.60     & 64.45    \\
		\bottomrule
	\end{tabular}
\end{table*}
\subsection{Evaluation on Cross-domain Attack}
In this part, we explore the effectiveness of cross-domain attacks, where the perturbation is generated from models trained on Waymo dataset \textit{without} having knowledge of data distribution of the target domain (KITTI dataset).
Based on our findings presented in Table~\ref{table:cross_domain_trans}, we gleaned several key insights:

\begin{enumerate}[1)]
	\item 
	\textit{Trans-domain Transferability}: Adversarial attacks exhibit \textit{transferability} across disparate domains. However, in most cases, cross-domain attacks do not exceed the effectiveness of source attacks. This observation may imply that the efficacy of adversarial attack techniques is somewhat tied to the data distribution of their originating domain.
	
	\item \textit{Exceptions in Cross-domain Attack Potency:} Interestingly, certain models, such as Voxel-RCNN and Part-A\^{}2, break from the general trend, showing higher susceptibility to cross-domain attacks than to source attacks. This unexpected behavior suggests that these models may have inherent vulnerabilities, which can be exploited by adversarial techniques trained on unrelated datasets.
	
	\item \textit{Vulnerability of Point-based Detectors}: Point-based detectors, particularly 3DSSD and PointRCNN, consistently exhibit vulnerability, even under cross-domain perturbations. Their susceptibility, evident in both source and cross-domain attacks, hints at potential structural weaknesses within point-based detectors.
\end{enumerate}

\section{Evaluation of General Defenses}\label{sec:eval_defense}

\subsection{Robustness Conferred by General Defenses}\label{sec:robustness_confer}
In this section, we examine the effectiveness of 8 defense methods against three types of attacks for 3D object detection. 
We apply simple transformations, including flipping, scaling, rotation, quantification, and random sampling, to adversarial examples before they are passed through the network.

Based on the results presented in Table~\ref{table:perturbation_defense}, we can derive several key insights regarding the robustness conferred by different defense methods against point perturbation attacks for various 3D object detection models.

\begin{enumerate}[1)]
	\item 
	Simple transformations, such as flipping, scaling, and rotation, bring substantial benefits, demonstrating their ability to mitigate the unnatural patterns imposed by adversarial attacks on the point cloud.
	Reviewing the average mAP ratios reveals that the rotation strategy is the most efficacious on average.
	The defense of rotation transformation appears to be exceptionally potent for several models, including SECOND, Voxel-RCNN, PointPillar, and Part-A\^{}2. 
	
	\item 
	The efficacy of defenses is not consistent across all models. For instance, while the rotation strategy works optimally for some models, others like PDV, 3DSSD, and PointRCNN benefit more from the flip defense. The absence of a universally potent defense method accentuates the need for model-agnostic defense approaches.
	
	\item 
	Although PA-AUG has been shown to enhance the robustness of detectors to corrupted data~\citep{choi2021part}, its effectiveness against adversarial attacks is limited, potentially due to differences in the distribution of corrupted data and adversarial examples.
	Defenses like Gaussian and KNN also tend to provide limited robustness enhancement for certain models. This could indicate that these strategies might not sufficiently disrupt adversarial patterns in the input data, and even bring more serious disturbances.
\end{enumerate}

\setlength{\tabcolsep}{8pt}
\begin{table*}[htbp]
	\centering
	\caption{Robustness of detectors (mAP ratio) under PGD-based \textbf{point perturbation attack} ($\epsilon$=10cm) after applying 9 general defense methods. For each detector, the result of the most effective defense method is bold. ``Gaussian": add Gaussian noises to the coordinates of points. ``Sample": random sampling. ``KNN": KNN-based denoising.}
	\label{table:perturbation_defense}
	\begin{tabular}{c|ccccccccc|c}
		\toprule
		\multirow{2}{*}{Model} & \multicolumn{9}{c|}{Defense}   & \multirow{2}{*}{Average}  \\
		& None  & Flip  & Scale & Rotate & Quantify & Gaussian & PA-AUG & Sample & KNN   \\ 
		\midrule
		SECOND  & 70.80 & 73.22 & 73.97 & \textbf{75.47}  & 69.68    & 69.42    & 72.54  & 71.98  & 64.76 & 71.32      \\
		Voxel-RCNN  & 80.24 & 82.73 & 82.10 & \textbf{84.75}  & 83.79    & 78.95    & 79.31  & 84.22  & 79.61 & 81.74      \\
		PDV     & 61.61 & \textbf{68.54} & 63.09 & 66.58  & 62.96    & 63.44    & 61.84  & 62.37  & 58.95 & 63.26      \\
		PointPillar  & 7.28  & 41.73 & 52.13 & \textbf{59.52}  & 11.53    & 14.08    & 9.79   & 9.46   & 9.38  & 23.88      \\
		Part-A\^{}2 & 75.11 & 80.01 & 78.22 & \textbf{81.57}  & 74.97    & 78.22    & 73.83  & 78.35  & 75.46 & 77.30      \\
		PV-RCNN & 49.80 & 51.64 & 48.38 & 54.11  & 47.87    & 45.58    & 50.26  & \textbf{56.43}  & 50.82 & 50.54      \\
		3DSSD   & 19.07 & \textbf{33.00} & 19.38 & 31.56  & 17.00    & 22.17    & 19.69  & 24.84  & 24.98 & 23.52      \\
		PointRCNN  & 14.44 & \textbf{28.21} & 15.86 & 25.44  & 16.46    & 16.04    & 15.78  & 25.84  & 18.04 & 19.57      \\ 
		\midrule
		Average & 47.29 & 57.38 & 54.14 & \textbf{59.88}  & 48.03    & 48.49    & 47.88  & 51.69  & 47.75 & 51.39                     \\
		\bottomrule
	\end{tabular}
\end{table*}

\setlength{\tabcolsep}{8pt}
\begin{table*}[htbp]
	\centering
	\caption{Robustness of detectors (mAP ratio) under \textbf{point detachment attack} after applying 9 defense methods. The ratio of detached points is set to 2\%. For each detector, the result of the most effective defense method is bold. ``Gaussian": add Gaussian noises to the coordinates of points. ``Sample": random sampling. ``KNN": KNN-based denoising.}
	\label{table:detachment_defense}
	\begin{tabular}{c|ccccccccc|c} 
		\toprule
		\multirow{2}{*}{Model} & \multicolumn{9}{c|}{Defense}      & \multirow{2}{*}{Average}  \\
		& None  & Flip           & Scale & Rotate         & Quantify & Gaussian & PA-AUG         & Sample & KNN   &                           \\ 
		\midrule
		SECOND                 & 67.92 & 66.88          & 69.33 & \textbf{71.05}          & 69.00    & 68.29    & 69.01          & 68.16  & 64.73 & 68.26                     \\
		Voxel-RCNN             & 64.84 & \textbf{66.74} & 65.58 & 66.18          & 66.17    & 65.51    & 64.38          & 65.75  & 63.73 & 65.43                     \\
		PDV                    & 48.13 & 48.29          & 49.19 & 48.96          & 48.76    & 49.34    &\textbf{ 49.71}          & 47.31  & 45.84 & 48.39                     \\
		PointPillar            & 65.14 & 62.72          & 65.15 & 64.47          & 65.01    & 64.46    & \textbf{66.70} & 63.75  & 63.04 & 64.49                     \\
		Part-A\^{}2              & 64.54 & 63.89          & 66.52 & 65.52          & 65.11    & 64.55    & \textbf{68.74}          & 65.93  & 62.94 & 65.30                     \\
		PV-RCNN                & 61.35 & \textbf{63.74}          & 63.20 & 62.89          & 61.66    & 59.77    & 62.98          & 59.77  & 60.65 & 61.78                     \\
		3DSSD                  & 57.75 & 56.73          & 57.19 & \textbf{57.47} & 56.04    & 57.19    & 56.01          & 57.19  & 56.01 & 56.84                     \\
		PointRCNN              & 68.49 & 68.55          & 70.09 & 67.71          & \textbf{71.63}    & 68.76    & 70.35          & 67.98  & 69.19 & 69.20                     \\ 
		\midrule
		Average                & 62.27 & 62.19          & 63.28 & 63.03          & 62.92    & 62.23    & \textbf{63.49}          & 61.98  & 60.77 & 62.46                     \\
		\bottomrule
	\end{tabular}
\end{table*}

\setlength{\tabcolsep}{8pt}
\begin{table*}[htbp]
	\centering
	\caption{Robustness of detectors (mAP ratio) under \textbf{point attachment attack} after applying 9 defense methods. For each detector, the result of the most effective defense method is bold. The number of attached points is set to 164 on the KITTI dataset. ``Gaussian": add Gaussian noises to the coordinates of points. ``Sample": random sampling. ``KNN": KNN-based denoising.}
	\label{table:attachment_defense}
	\begin{tabular}{c|ccccccccc|c} 
		\toprule
		\multirow{2}{*}{Model} & \multicolumn{9}{c|}{Defense}     & \multirow{2}{*}{Average}  \\
		& None  & Flip           & Scale          & Rotate & Quantify & Gaussian & PA-AUG         & Sample & KNN   &                           \\ 
		\midrule
		SECOND                 & 85.50 & \textbf{92.40} & 91.29          & 88.51  & 85.76    & 85.18    & 86.81          & 85.09  & 84.19 & 87.19                     \\
		Voxel-RCNN             & 91.45 & 88.18          & 90.40          & 88.84  & 89.30    & 89.26    & \textbf{92.01} & 88.42  & 87.59 & 89.50                     \\
		PDV                    & 64.14 & \textbf{78.56} & 68.86          & 68.80  & 64.91    & 66.35    & 66.57          & 68.16  & 67.28 & 68.18                     \\
		PointPillar            & 89.80 & 87.03          & \textbf{93.89} & 92.45  & 92.38    & 90.80    & 89.95          & 90.03  & 88.21 & 90.51                     \\
		Part-A\^{}2            & 92.58 & 86.81          & \textbf{94.90}          & 93.56  & 93.52    & 91.67    & 93.41          & 94.70  & 92.93 & 92.68                     \\
		PV-RCNN                & 90.56 & 87.88          & 91.58          & \textbf{91.82}  & 87.46    & 89.47    & 89.76          & 90.50  & 86.70 & 89.52                     \\
		3DSSD                  & 68.27 & \textbf{84.34} & 77.14          & 77.85  & 75.05    & 75.30    & 70.48          & 77.09  & 75.58 & 75.68                     \\
		PointRCNN              & 72.86 & 85.49          & \textbf{92.54} & 89.01  & 90.41    & 88.79    & 76.68          & 86.99  & 86.24 & 85.45                     \\ 
		\midrule
		Average                & 81.89 & 86.33          & \textbf{87.58}          & 86.36  & 84.85    & 84.60    & 82.83          & 85.12  & 83.59 & 84.84                     \\
		\bottomrule
	\end{tabular}
\end{table*}


\begin{table*}
	\centering
	\caption{
		Robustness of detectors (mAP ratio) after applying 7 transformation-based defense methods under \textit{\textbf{adaptive}} PGD-based point perturbation attack ($\epsilon$=10cm). For each detector, the result of the most effective defense method is bold.}
	\label{table:adaptive_perturbation_defense}
	\begin{tabular}{c|cccccccc} 
		\toprule
		\diagbox{Method}{Defense} & None  & Flip  & Scale & Rotate & Quantify & Gaussian & Sample & KNN    \\ 
		\hline
		SECOND      & 70.80 & 68.01 & 68.77 & \textbf{70.56}  & 68.63    & 65.53    & 66.57  & 61.43  \\
		Voxel-RCNN  & 80.24 & 80.82 & 80.41 & 80.92  & 81.27    & 79.69    & \textbf{81.16}  & 77.98  \\
		PDV         & 61.61 & 62.06 & \textbf{62.49} & 61.77  & 61.82    & 58.17    & 62.48  & 57.02  \\
		PointPillar & 7.28  & \textbf{8.62}  & 7.71  & 7.66   & 7.75     & 6.83     & 7.51   & 7.08   \\
		Part-A\^{}2    & 75.11 & 77.55 & 74.37 & 75.61  & 76.80    & 75.80    & 77.10  & \textbf{77.75}  \\
		PV-RCNN     & 49.80 & 45.83 & \textbf{51.18} & 49.57  & 47.74    & 47.93    & 50.80  & 43.46  \\
		3DSSD       & 19.07 & 16.46 & 18.89 & 19.44  & \textbf{19.65}    & 18.67    & 19.40  & 14.80  \\
		PointRCNN   & 14.44 & \textbf{17.72} & 16.23 & 16.83  & 17.04    & 14.35    & 14.85  & 14.13  \\
		\bottomrule
	\end{tabular}
\end{table*}

Table~\ref{table:detachment_defense} shows the mAP ratios of detectors under point detachment attack when different defense methods are applied.  And we have several observations:
\begin{enumerate}[1)]

\item Looking at the average mAP ratios, while the differences are slight, the PA-AUG defense emerges as the most effective on average. This implies that appropriate data augmentation might offer a generic robustness enhancement against point detachment. 

\item Neither KNN-based denoising nor random sampling enhances the average mAP ratios of detectors under point detachment attacks because they reduce the number of points. Unlike the point perturbation attack, the point detachment attack does not impose perturbation on point clouds, so removing points in the two defense methods is not helpful to mitigate the imposed perturbation.
\end{enumerate}

The results presented in Table~\ref{table:attachment_defense} illustrate the effectiveness of different defense methods against point attachment attacks on various 3D object detection models.
Simple transformations, like flipping, scaling, and rotation, are superior in countering the disturbances caused by attached points.
Unexpectedly, KNN-based denoising generally does not appear to be an effective defense method, indicating its limitations in dealing with attached points by removing the outlier of clusters.

\subsection{Defense Against Adaptive Attacks}
As discussed in Section~\ref{sec:robustness_confer}, simple transformations including flipping, rotation, scaling, quantification, and random sampling can make the powerful PGD-based point perturbation attack less malignant. However, these transformations are applied post-generation of the adversarial examples. So defenses in this way assume that knowledge of used defense mechanisms is \textbf{not} exploited when generating adversarial examples. 
According to Kerckhoff's principle~\citep{Kerckhoffs1883}, a system should be safe even if all information about it except for the key is publicly available knowledge. So in this section, we further study whether the defense is effective against \textbf{adaptive attacks}, which generated the adversarial examples based on the transformed point cloud input.

Specifically, we use the transformed point cloud as input for generating adversarial examples. Particularly, for the defense of adding Gaussian noises, we apply Expectation Over Transformation (EOT)~\citep{athalye18a} to modify the iterative FGM updation (see Eq.~\ref{eq:calculate_gradient}) to compute the expected gradient over the distribution of imposed Gaussian noises to the input:
\begin{equation}\label{eq:adaptive_adversarial_attack_iteration} 
	\mathbf{X}_a^{t+1} = \texttt{Clip}_{\mathbf{X},\epsilon}\left\{  \mathbf{X}_a^t - \alpha \mathbb{E}_{t\sim \mathcal{T}}  \frac{ \bigtriangledown_{\mathbf{X}_a} J(t(\mathbf{X}_a), \mathcal{G}) }{\|  \bigtriangledown_{\mathbf{X}_a} J(t(\mathbf{X}_a), \mathcal{G})  \|_2}\right\}, 
\end{equation}
where $\mathcal{T}$ is the distribution of imposed Gaussian noises $t$. 
Specifically, we compute the average of normalized gradients from multiple transformed point clouds and use this average to update the perturbation in each iteration

We report the detailed results of (PGD-based) adaptive point perturbation attack on the KITTI dataset in Table~\ref{table:adaptive_perturbation_defense}.
We can observe none of the various transformations could provide a substantial robustness boost against adaptive attacks, as the mAP ratio of detectors with defense is similar to the value under setting without defense. 
For instance, the rotation transformation could elevate the mAP ratio of PointPillar from 7.28\% to 59.52\% under a standard attack (as shown in Table~\ref{table:perturbation_defense}), but it only marginally increases the mAP ratio to 7.66\%, as seen in Table~\ref{table:adaptive_perturbation_defense}.
\textit{The experimental results suggest that adversarial attacks are not affected by input transformation defense if the adopted transformation is known to the attacker.}

\color{black}
\noindent\textbf{Discussion.}
The inconsistent performance of various defense strategies, coupled with their lackluster response to adaptive attacks, underscores the pressing need for a universally effective defense mechanism.
Adversarial training offers a promising avenue in this context, which aims to cultivate models that are intrinsically robust rather than modifying the input point cloud. 
In the next section, we validate the effectiveness of our proposed balanced adversarial focal training.
\color{black}

\color{black}
\section{Evaluation of BAFT}\label{sec:eval_baft}
In this part, we conduct experiments to compare the results of our proposed \textit{balanced adversarial focal training} and conventional adversarial training on clean and adversarial examples. Besides, we also evaluate the robustness of the model trained with BAFT on the corrupted data.
\subsection{Comparison with Conventional Adversarial Training}

\setlength{\tabcolsep}{4pt}
\begin{table*}[htp]
	\centering
	\caption{
		\revise{Comparison of different training strategies on the KITTI dataset. The mAP of various detectors on clean and adversarial examples is reported under three different types of attacks. ``Normal" denotes standard training, ``AT" is conventional adversarial training.}
	}
	\label{table:BAFT_eval}
	\begin{tabular}{c|c|c|cccccccc} 
		\toprule
		Attack Type                                                                                          & Validation                   & Setting     & SECOND & Voxel-RCNN & PDV   & PointPillar & Part-A\^{}2 & PV-RCNN & 3DSSD & PointRCNN  \\ 
		\hline
		\multirow{6}{*}{\begin{tabular}[c]{@{}c@{}}Point \\ Perturbation \\ $\epsilon$ of 10cm\end{tabular}} & \multirow{3}{*}{Clean}       & Normal      & 69.64  & 74.36      & 75.91 & 67.22       & 74.85    & 71.03   & 71.48 & 72.62      \\
		&                              & AT          & 65.62  & 73.19      & 71.65 & 64.69       & 72.68    & 73.31   & 69.90 & 70.94      \\
		&                              & BAFT (Ours) & 70.35  & 76.61      & 74.34 & 65.16       & 74.12    & 75.23   & 71.33 & 72.27      \\ 
		\cline{2-11}
		& \multirow{3}{*}{Adversarial} & Normal      & 49.30  & 59.67      & 46.77 & 4.90        & 56.22    & 35.37   & 13.63 & 10.49      \\
		&                              & AT          & 84.99  & 69.71      & 87.61 & 47.16       & 83.29    & 60.53   & 34.67 & 38.36      \\
		&                              & BAFT (Ours) & 77.27  & 78.42      & 76.61 & 44.94       & 82.11    & 67.48   & 31.81 & 35.24      \\ 
		\hline
		\multirow{6}{*}{\begin{tabular}[c]{@{}c@{}}Point \\ Detachment \\ 2\% points\end{tabular}}           & \multirow{3}{*}{Clean}       & Normal      & 69.64  & 74.36      & 75.91 & 67.22       & 74.85    & 71.03   & 71.48 & 72.62      \\
		&                              & AT          & 69.34 &	75.78 &	75.23 &	67.17 &	75.37 &	74.95 &	71.47 &	72.27 \\
		&                              & BAFT (Ours) & 70.34  & 76.68      & 76.54 & 69.43       & 77.43    & 75.81   & 72.50 & 75.64      \\ 
		\cline{2-11}
		& \multirow{3}{*}{Adversarial} & Normal      & 47.30  & 48.21      & 36.54 & 43.79       & 48.31    & 44.90   & 41.28 & 49.74      \\
		&                              & AT          & 56.80  & 52.26      & 58.14 & 45.86       & 58.44    & 45.68   & 40.08 & 52.69      \\
		&                              & BAFT (Ours) & 51.32  & 51.82      & 60.82 & 50.37       & 54.33    & 46.41   & 41.20 & 53.80      \\ 
		\hline
		\multirow{6}{*}{\begin{tabular}[c]{@{}c@{}}Point \\ Attachment \\ 164 points\end{tabular}}           & \multirow{3}{*}{Clean}       & Normal      & 69.64  & 74.36      & 75.91 & 67.22       & 74.85    & 71.03   & 71.48 & 72.62      \\
		&                              & AT          & 69.26  & 74.51      & 74.25 & 65.74       & 73.63    & 72.94   & 36.00 & 36.93      \\
		&                              & BAFT (Ours) & 69.92  & 75.27      & 75.37 & 66.36       & 74.39    & 73.36   & 54.67 & 63.14      \\ 
		\cline{2-11}
		& \multirow{3}{*}{Adversarial} & Normal      & 60.40  & 66.82      & 58.49 & 55.39       & 71.47    & 62.52   & 59.23 & 56.80      \\
		&                              & AT          & 66.67  & 75.19      & 76.48 & 58.99       & 74.17    & 64.55   & 35.96 & 30.35      \\
		&                              & BAFT (Ours) & 64.61  & 73.13      & 76.85 & 60.29       & 73.31    & 63.54   & 53.54 & 54.54      \\
		\bottomrule
	\end{tabular}
\end{table*}
\setlength{\tabcolsep}{6pt}

We provide a comprehensive comparison between the performance of various detectors using different training strategies on both clean and adversarial examples, as presented in Table~\ref{table:BAFT_eval}. There are several critical observations:
\begin{enumerate}[1)]
	\item Generally, AT shows a remarkable improvement in adversarial robustness. However, for some detectors, its performance on clean data is compromised.
	Adversarial training, while effective at improving the robustness against adversarial attacks (such as the point perturbation attack where the mAP of SECOND increased from 49.3\% to 84.99\%), reveals a drawback. It often compromises the performance on clean (non-adversarial) examples, as evidenced by the performance drop of the SECOND detector from 69.64\% to 65.62\%.
	
	\item The results showcase that BAFT seems to address the trade-off presented by the AT method. With BAFT, not only does the adversarial robustness remain commendable, but there is also an improvement in the performance on clean examples. The SECOND detector, for example, improved on clean examples by 0.71\% (from 69.64\% to 70.35\%), and yet maintained an mAP of 77.27\% on adversarial examples. This demonstrates that BAFT potentially prevents overfitting to adversarial examples and retains, or even enhances, the performance on clean examples.
	
	\item This effectiveness of BAFT is not limited to just one type of attack. Under various attack scenarios like point detachment and point attachment, similar phenomena are observed. For instance, with the PointRCNN detector under point attachment, the AT significantly reduced the performance on clean examples (from 72.62\% to 36.93\%), while BAFT retained a reasonable performance of 63.14\% mAP.
	
	\item The ability of BAFT to improve the performance of 3D detectors on clean examples suggests its potential application beyond just adversarial robustness. It can be used as a data augmentation strategy to enhance general performance.
\end{enumerate}


In essence, the proposed BAFT not only seems to strike a balance between adversarial robustness and clean data performance but, in some cases, even offers dual benefits by potentially serving as a data augmentation technique.

\subsection{Ablation Study of Focal Weights}
\begin{table}[t]
	\centering
	\caption{\revise{Comparison of different training strategies for PDV. ``Normal" denotes standard training, ``AT" is conventional adversarial training, ``ADD" means directly adding the loss on clean and adversarial examples together. We report the mAP of detectors on clean examples and adversarial examples under PGD-based perturbation attack with $\epsilon$ of 10cm on the KITTI dataset.}}
	\label{table:ablation_focal_weights}
	\begin{tabular}{c|cc} 
		\toprule
		Setting & Clean Examples    & Adversarial Examples      \\ 
		\hline
		Nomral  & 75.91 & 46.77  \\
		AT    & 71.65 & 87.61  \\
		ADD     & 73.59 & 72.48       \\
		BAFT (Ours)    & \textbf{74.34} & \textbf{76.61}  \\
		\bottomrule
	\end{tabular}
\end{table}


Table~\ref{table:ablation_focal_weights} presents a quantitative evaluation of PDV using various training strategies. Specifically, the performance of the model is assessed on both clean examples and adversarial examples subjected to a PGD-based perturbation attack with an $\epsilon$ of 10cm on the KITTI dataset.
The ``Add" strategy, which simply combines the loss from clean and adversarial examples, leads to a noticeable improvement of 1.94\% in mAP on clean examples. However, it comes at the expense of a significant 15.13\% decrease in mAP on adversarial examples compared to conventional adversarial training.
Remarkably, our proposed BAFT strategy surpasses the baseline ``Add" strategy on both fronts. It achieves a competitive mAP of 74.34\% on clean examples, demonstrating a minimal drop in clean example accuracy while effectively enhancing the robustness of models. On adversarial examples, BAFT achieves an impressive mAP of 76.61\%, showcasing its ability to counter adversarial attacks effectively.
Notably, the dynamic adjustment of loss weights in BAFT yields a 0.75\% improvement on clean examples and a substantial 4.13\% increase in mAP on adversarial examples when compared to the ``Add" strategy. These results validate the effectiveness of our proposed technique in aligning and harmonizing diverse loss components, thus effectively balancing the robustness and accuracy in 3D object detection.

\subsection{Evaluation on the Corrupted Examples}\label{sec:evaluation_corruption}
\begin{figure*}[htp]
	\centering
		\includegraphics[width=\textwidth]{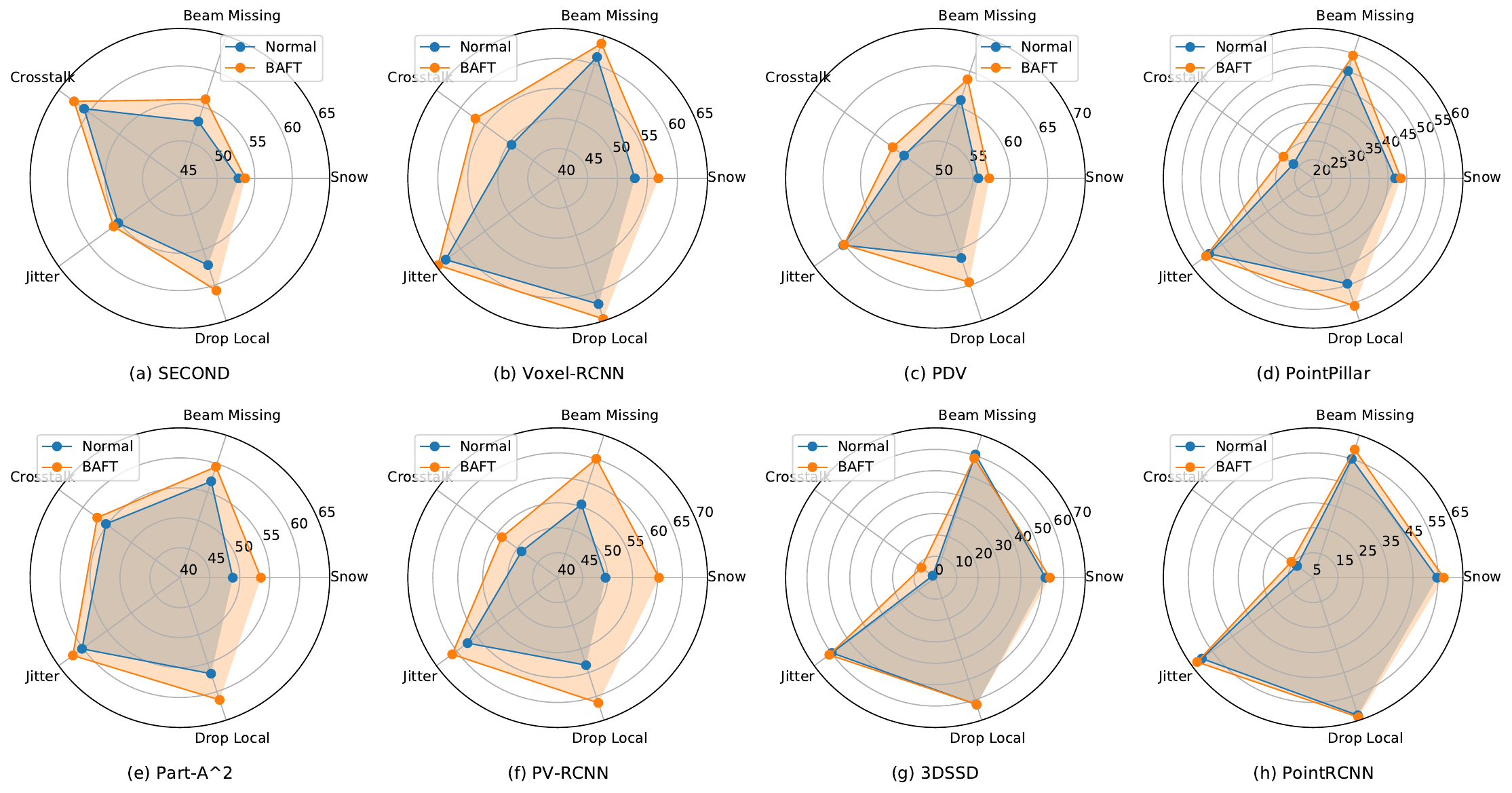} 
		\caption{
			\revise{Comparison of detectors with different training strategies on corrupted data from the KITTI dataset. Each radar plot illustrates the performance difference between a detector trained in the conventional way and one trained with the Balanced Adversarial Focal Training (BAFT) strategy. The mAP metric is used for this comparison.}
		}
		\label{fig:perturbation_radar}
\end{figure*}
This section offers a detailed analysis of the impact of the proposed BAFT when faced with varied corruptions in the KITTI dataset. By contrasting the results from normal training with those from our Balanced Adversarial Focal Training (BAFT) strategy, we intend to emphasize the effectiveness of BAFT in bolstering detector robustness in the face of corrupted data.

The radar plots depicted in Fig.~\ref{fig:perturbation_radar} present the mAP performance of detectors against a range of corruptions, namely: beam missing, snow, drop local, jitter, and crosstalk. For a clear understanding of the corruption scenarios applied in this analysis, refer to section \ref{sec:implementation_details} for implementation details. 
The radar plot clearly indicates that the detectors using BAFT consistently outperform their counterparts using the normal training strategy, validating the ability of BAFT to elevate the robustness of detectors against diverse corruptions. Taking the SECOND as an example, BAFT with adversarial examples generated by point detachment attack improves the mAP for beam missing, drop local, and snow corruption by 3.12\%, 3.58\%, and 0.88\%, respectively.

In summary, the comparison of detectors under normal training and BAFT highlights the substantial improvements in corruption robustness achieved through our proposed strategy. 
The adversarial training can be regarded as a data augmentation strategy that trains models with adversarial examples. 
Adversarial examples, created by introducing unnatural patterns into point clouds, remain \textit{plausible} even in the absence of a malicious attacker. For instance, real-world corruption can arise from various scenarios in autonomous driving. Hence, the capability of BAFT to enhance performance on corrupted data aligns with real-world challenges encountered in autonomous driving scenarios, where such corruption is naturally present.

\section{Discussion and Conclusion}\label{sec:conclusion}

In this paper, we thoroughly investigated the robustness of state-of-the-art LiDAR-based 3D object detectors under three basic adversarial attacks. We carefully analyzed the relationship between the robustness and the properties of detectors in detail. We also demonstrated the transferability of cross-model, cross-task, and cross-domain attacks. Furthermore, we evaluated the effectiveness of general defense strategies against adversarial attacks and the proposed BAFT. After conducting thorough experiments, we arrived at several significant insights, as summarized below:

\begin{enumerate}[1)]
	\item 
	While the point perturbation attack is generally stronger than the other two types of attacks, its adversarial examples remain subtly perceptible.
	
	\item When comparing robustness to the potent point perturbation attack, voxel-based detectors generally exhibit greater robustness than point-based detectors. This is attributed to certain inherent challenges the point perturbation attack faces when targeting voxel-based detectors.
	
	\item Factors like model size, whether the model is of a single-stage or two-stage design, and accuracy on clean examples have minimal impact on the robustness of detectors. Notably, integrating local features into pillar feature encoding significantly reduces the robustness of PointPillar against the point perturbation attack.
	
	\item Adversarial examples derived from more vulnerable point-based detectors, notably PointRCNN and 3DSSD, also prove effective at compromising other detectors. The clear transferability seen in cross-model and cross-task attacks offers insights into developing stronger and more adaptable attacks.
	
	\item Relying solely on simple transformation as a defense strategy offers minimal enhancement in robustness when the specifics of the defense strategy are public knowledge for the attacker.
	
	\item While adversarial training shows promise in augmenting the robustness of 3D detectors, it inadvertently undermines the accuracy of models on clean examples. Our proposed BAFT addresses this trade-off, offering a more balanced solution.
	
\end{enumerate}

We firmly believe that our extensive experiments and meticulous observations will greatly contribute to advancing the robustness and reliability of LiDAR-based 3D object detectors, thus paving the way for future endeavors in this field.

\color{black}

\section*{Data Availability Statements}
The Waymo Open Dataset~\citep{Sun_2020_CVPR} and KITTI~\citep{Geiger_KITTI} used in this manuscript are deposited in publicly available repositories respectively: \url{https://waymo.com/open/data/perception} and \url{http://www.cvlibs.net/datasets/kitti}.

\section*{Conflict of Interest}
The authors declare that they do not have any commercial or associative interest that represents a conflict of interest in connection with the work submitted.

\bibliographystyle{spbasic}
\bibliography{bib}

\begin{thebibliography}{64}
\providecommand{\natexlab}[1]{#1}
\providecommand{\url}[1]{{#1}}
\providecommand{\urlprefix}{URL }
\expandafter\ifx\csname urlstyle\endcsname\relax
  \providecommand{\doi}[1]{DOI~\discretionary{}{}{}#1}\else
  \providecommand{\doi}{DOI~\discretionary{}{}{}\begingroup
  \urlstyle{rm}\Url}\fi
\providecommand{\eprint}[2][]{\url{#2}}

\bibitem[{Abdelfattah et~al.(2021)Abdelfattah, Yuan, Wang, and
  Ward}]{abdelfattah2021adversarial}
Abdelfattah M, Yuan K, Wang ZJ, Ward R (2021) Adversarial attacks on
  camera-lidar models for 3d car detection. In: 2021 IEEE/RSJ International
  Conference on Intelligent Robots and Systems (IROS), IEEE, pp 2189--2194

\bibitem[{Athalye et~al.(2018)Athalye, Carlini, and Wagner}]{athalye18a}
Athalye A, Carlini N, Wagner D (2018) Obfuscated gradients give a false sense
  of security: Circumventing defenses to adversarial examples. In:
  International Conference on Machine Learning, vol~80, pp 274--283

\bibitem[{Bri{\~n}{\'o}n-Arranz et~al.(2021)Bri{\~n}{\'o}n-Arranz, Rakotovao,
  Creuzet, Karaoguz, and El-Hamzaoui}]{brinon2021methodology}
Bri{\~n}{\'o}n-Arranz L, Rakotovao T, Creuzet T, Karaoguz C, El-Hamzaoui O
  (2021) A methodology for analyzing the impact of crosstalk on lidar
  measurements. In: Sensors, IEEE, pp 1--4

\bibitem[{Carlini and Wagner(2017)}]{carlini2017towards}
Carlini N, Wagner D (2017) Towards evaluating the robustness of neural
  networks. In: IEEE Symposium on Security and Privacy, Ieee, pp 39--57

\bibitem[{Carrilho et~al.(2018)Carrilho, Galo, and
  Dos~Santos}]{carrilho2018statistical}
Carrilho A, Galo M, Dos~Santos RC (2018) Statistical outlier detection method
  for airborne lidar data. International Archives of the Photogrammetry, Remote
  Sensing \& Spatial Information Sciences 42(1)

\bibitem[{Chen et~al.(2022)Chen, Chen, Zhang, and Tao}]{chen2022sasa}
Chen C, Chen Z, Zhang J, Tao D (2022) Sasa: Semantics-augmented set abstraction
  for point-based 3d object detection. In: Proceedings of the AAAI Conference
  on Artificial Intelligence, vol~1, pp 221--229

\bibitem[{Cheng et~al.(2022)Cheng, Liang, Choi, Tao, Cao, Liu, and
  Zhang}]{cheng2022physical}
Cheng Z, Liang J, Choi H, Tao G, Cao Z, Liu D, Zhang X (2022) Physical attack
  on monocular depth estimation with optimal adversarial patches. In: European
  Conference on Computer Vision, Springer, pp 514--532

\bibitem[{Choi et~al.(2021)Choi, Song, and Kwak}]{choi2021part}
Choi J, Song Y, Kwak N (2021) Part-aware data augmentation for 3d object
  detection in point cloud. In: IEEE/RSJ International Conference on
  Intelligent Robots and Systems, pp 3391--3397

\bibitem[{Deng et~al.(2021)Deng, Shi, Li, Zhou, Zhang, and Li}]{deng2021voxel}
Deng J, Shi S, Li P, Zhou W, Zhang Y, Li H (2021) Voxel r-cnn: Towards high
  performance voxel-based 3d object detection. In: Proceedings of the AAAI
  Conference on Artificial Intelligence, pp 1201--1209

\bibitem[{Dong et~al.(2020)Dong, Chen, Zhou, Hua, Zhang, and Yu}]{dong2020self}
Dong X, Chen D, Zhou H, Hua G, Zhang W, Yu N (2020) Self-robust 3d point
  recognition via gather-vector guidance. In: Proceedings of the IEEE/CVF
  Conference on Computer Vision and Pattern Recognition, IEEE, pp 11513--11521

\bibitem[{Dong et~al.(2018)Dong, Liao, Pang, Su, Zhu, Hu, and
  Li}]{dong2018boosting}
Dong Y, Liao F, Pang T, Su H, Zhu J, Hu X, Li J (2018) Boosting adversarial
  attacks with momentum. In: Proceedings of the IEEE/CVF Conference on Computer
  Vision and Pattern Recognition, pp 9185--9193

\bibitem[{Dziugaite et~al.(2016)Dziugaite, Ghahramani, and
  Roy}]{dziugaite2016study}
Dziugaite GK, Ghahramani Z, Roy DM (2016) A study of the effect of jpg
  compression on adversarial images. arXiv preprint arXiv:160800853

\bibitem[{Geiger et~al.(2012)Geiger, Lenz, and Urtasun}]{Geiger_KITTI}
Geiger A, Lenz P, Urtasun R (2012) Are we ready for autonomous driving? the
  kitti vision benchmark suite. In: Proceedings of the IEEE/CVF Conference on
  Computer Vision and Pattern Recognition, pp 3354--3361

\bibitem[{Goodfellow et~al.(2014)Goodfellow, Shlens, and
  Szegedy}]{goodfellow2014explaining}
Goodfellow IJ, Shlens J, Szegedy C (2014) Explaining and harnessing adversarial
  examples. arXiv preprint arXiv:14126572

\bibitem[{Guo et~al.(2018)Guo, Rana, Cisse, and van~der
  Maaten}]{guo2018countering}
Guo C, Rana M, Cisse M, van~der Maaten L (2018) Countering adversarial images
  using input transformations. In: International Conference on Learning
  Representations

\bibitem[{Hahner et~al.(2022)Hahner, Sakaridis, Bijelic, Heide, Yu, Dai, and
  Van~Gool}]{hahner2022lidar}
Hahner M, Sakaridis C, Bijelic M, Heide F, Yu F, Dai D, Van~Gool L (2022) Lidar
  snowfall simulation for robust 3d object detection. In: Proceedings of the
  IEEE/CVF Conference on Computer Vision and Pattern Recognition, pp
  16364--16374

\bibitem[{Hamdi et~al.(2020)Hamdi, Rojas, Thabet, and Ghanem}]{hamdi2020advpc}
Hamdi A, Rojas S, Thabet A, Ghanem B (2020) Advpc: Transferable adversarial
  perturbations on 3d point clouds. In: European Conference on Computer Vision,
  Springer, pp 241--257

\bibitem[{He et~al.(2020)He, Zeng, Huang, Hua, and Zhang}]{He_2020_CVPR}
He C, Zeng H, Huang J, Hua XS, Zhang L (2020) Structure aware single-stage 3d
  object detection from point cloud. In: Proceedings of the IEEE/CVF Conference
  on Computer Vision and Pattern Recognition, pp 11870--11879

\bibitem[{Herrmann et~al.(2022)Herrmann, Sargent, Jiang, Zabih, Chang, Liu,
  Krishnan, and Sun}]{Herrmann_2022_CVPR}
Herrmann C, Sargent K, Jiang L, Zabih R, Chang H, Liu C, Krishnan D, Sun D
  (2022) Pyramid adversarial training improves vit performance. In: Proceedings
  of the IEEE/CVF Conference on Computer Vision and Pattern Recognition, pp
  13419--13429

\bibitem[{Hu et~al.(2022{\natexlab{a}})Hu, Kuai, and Waslander}]{hu2022point}
Hu JS, Kuai T, Waslander SL (2022{\natexlab{a}}) Point density-aware voxels for
  lidar 3d object detection. In: Proceedings of the IEEE/CVF Conference on
  Computer Vision and Pattern Recognition, pp 8469--8478

\bibitem[{Hu et~al.(2023)Hu, Zhang, Liu, Hou, Li, Zhang, Jin, and
  Sun}]{hu2023pointca}
Hu S, Zhang J, Liu W, Hou J, Li M, Zhang LY, Jin H, Sun L (2023) Pointca:
  Evaluating the robustness of 3d point cloud completion models against
  adversarial examples. In: Proceedings of the AAAI Conference on Artificial
  Intelligence, vol~37, pp 872--880

\bibitem[{Hu et~al.(2022{\natexlab{b}})Hu, Huang, Zhu, Sun, Zhang, and
  Hu}]{hu2022adversarial}
Hu Z, Huang S, Zhu X, Sun F, Zhang B, Hu X (2022{\natexlab{b}}) Adversarial
  texture for fooling person detectors in the physical world. In: Proceedings
  of the IEEE/CVF Conference on Computer Vision and Pattern Recognition, pp
  13307--13316

\bibitem[{Jia et~al.(2021)Jia, Cao, and Gong}]{jia2021intrinsic}
Jia J, Cao X, Gong NZ (2021) Intrinsic certified robustness of bagging against
  data poisoning attacks. In: Proceedings of the AAAI Conference on Artificial
  Intelligence, vol~35, pp 7961--7969

\bibitem[{Kerckhoffs(1883)}]{Kerckhoffs1883}
Kerckhoffs A (1883) La cryptographie militaire. Journal des sciences militaires
  9:5--38

\bibitem[{Kong et~al.(2023)Kong, Liu, Li, Chen, Zhang, Ren, Pan, Chen, and
  Liu}]{kong2023robo3d}
Kong L, Liu Y, Li X, Chen R, Zhang W, Ren J, Pan L, Chen K, Liu Z (2023)
  Robo3d: Towards robust and reliable 3d perception against corruptions. arXiv
  preprint arXiv:230317597

\bibitem[{Kurakin et~al.(2017)Kurakin, Goodfellow, and
  Bengio}]{kurakin2017adversarial}
Kurakin A, Goodfellow IJ, Bengio S (2017) Adversarial machine learning at
  scale. In: International Conference on Learning Representations

\bibitem[{Kurakin et~al.(2018)Kurakin, Goodfellow, and
  Bengio}]{kurakin2018adversarial}
Kurakin A, Goodfellow IJ, Bengio S (2018) Adversarial examples in the physical
  world. In: Artificial intelligence safety and security, Chapman and Hall/CRC,
  pp 99--112

\bibitem[{Lang et~al.(2019{\natexlab{a}})Lang, Vora, Caesar, Zhou, Yang, and
  Beijbom}]{Lang_2019_CVPR}
Lang A, Vora S, Caesar H, Zhou L, Yang J, Beijbom O (2019{\natexlab{a}})
  Pointpillars: Fast encoders for object detection from point clouds. In:
  Proceedings of the IEEE/CVF Conference on Computer Vision and Pattern
  Recognition, pp 12689--12697

\bibitem[{Lang et~al.(2019{\natexlab{b}})Lang, Vora, Caesar, Zhou, Yang, and
  Beijbom}]{LangVCZYB19}
Lang AH, Vora S, Caesar H, Zhou L, Yang J, Beijbom O (2019{\natexlab{b}})
  Pointpillars: Fast encoders for object detection from point clouds. In:
  Proceedings of the IEEE/CVF Conference on Computer Vision and Pattern
  Recognition, pp 12697--12705

\bibitem[{Lehner et~al.(2022)Lehner, Gasperini, Marcos-Ramiro, Schmidt, Mahani,
  Navab, Busam, and Tombari}]{lehner20223d}
Lehner A, Gasperini S, Marcos-Ramiro A, Schmidt M, Mahani MAN, Navab N, Busam
  B, Tombari F (2022) 3d-vfield: Adversarial augmentation of point clouds for
  domain generalization in 3d object detection. In: Proceedings of the IEEE/CVF
  Conference on Computer Vision and Pattern Recognition, pp 17295--17304

\bibitem[{Li et~al.(2019)Li, Chen, Wang, and Carin}]{li2019certified}
Li B, Chen C, Wang W, Carin L (2019) Certified adversarial robustness with
  additive noise. Advances in neural information processing systems 32

\bibitem[{Lin et~al.(2017)Lin, Goyal, Girshick, He, and
  Doll{\'a}r}]{lin2017focal}
Lin TY, Goyal P, Girshick R, He K, Doll{\'a}r P (2017) Focal loss for dense
  object detection. In: Proceedings of the IEEE/CVF International Conference on
  Computer Vision, pp 2980--2988

\bibitem[{Liu and Hu(2022)}]{liu2022imperceptible}
Liu D, Hu W (2022) Imperceptible transfer attack and defense on 3d point cloud
  classification. IEEE Transactions on Pattern Analysis and Machine
  Intelligence

\bibitem[{Liu et~al.(2019)Liu, Yu, and Su}]{liu2019extending}
Liu D, Yu R, Su H (2019) Extending adversarial attacks and defenses to deep 3d
  point cloud classifiers. In: 2019 IEEE International Conference on Image
  Processing (ICIP), IEEE, pp 2279--2283

\bibitem[{Liu et~al.(2020)Liu, Yu, and Su}]{liu2020adversarial}
Liu D, Yu R, Su H (2020) Adversarial shape perturbations on 3d point clouds.
  In: European Conference on Computer Vision, Springer, pp 88--104

\bibitem[{Liu et~al.(2021)Liu, Jia, and Gong}]{Liu_2021_CVPR}
Liu H, Jia J, Gong NZ (2021) Pointguard: Provably robust 3d point cloud
  classification. In: Proceedings of the IEEE/CVF Conference on Computer Vision
  and Pattern Recognition, pp 6186--6195

\bibitem[{Madry et~al.(2018)Madry, Makelov, Schmidt, Tsipras, and
  Vladu}]{MadryMSTV18}
Madry A, Makelov A, Schmidt L, Tsipras D, Vladu A (2018) Towards deep learning
  models resistant to adversarial attacks. In: International Conference on
  Learning Representations, pp 1--28

\bibitem[{Osadchy et~al.(2017)Osadchy, Hernandez-Castro, Gibson, Dunkelman, and
  P{\'e}rez-Cabo}]{osadchy2017no}
Osadchy M, Hernandez-Castro J, Gibson S, Dunkelman O, P{\'e}rez-Cabo D (2017)
  No bot expects the deepcaptcha! introducing immutable adversarial examples,
  with applications to captcha generation. IEEE Transactions on Information
  Forensics and Security 12(11):2640--2653

\bibitem[{Paszke et~al.(2019)Paszke, Gross, Massa, Lerer, Bradbury, Chanan,
  Killeen, Lin, Gimelshein, Antiga, Desmaison, K{\"{o}}pf, Yang, DeVito,
  Raison, Tejani, Chilamkurthy, Steiner, Fang, Bai, and Chintala}]{Pytorch}
Paszke A, Gross S, Massa F, Lerer A, Bradbury J, Chanan G, Killeen T, Lin Z,
  Gimelshein N, Antiga L, Desmaison A, K{\"{o}}pf A, Yang EZ, DeVito Z, Raison
  M, Tejani A, Chilamkurthy S, Steiner B, Fang L, Bai J, Chintala S (2019)
  Pytorch: An imperative style, high-performance deep learning library. In:
  Wallach HM, Larochelle H, Beygelzimer A, d'Alch{\'{e}}{-}Buc F, Fox EB,
  Garnett R (eds) Advances in Neural Information Processing Systems, pp
  8024--8035

\bibitem[{Qi et~al.(2017)Qi, Su, Mo, and Guibas}]{qi2017pointnet}
Qi C, Su H, Mo K, Guibas LJ (2017) Pointnet: Deep learning on point sets for 3d
  classification and segmentation. In: Proceedings of the IEEE conference on
  computer vision and pattern recognition, pp 652--660

\bibitem[{Shi et~al.(2019)Shi, Wang, and Li}]{shi2019pointrcnn}
Shi S, Wang X, Li H (2019) Pointrcnn: 3d object proposal generation and
  detection from point cloud. In: Proceedings of the IEEE/CVF conference on
  computer vision and pattern recognition, pp 770--779

\bibitem[{Shi et~al.(2020{\natexlab{a}})Shi, Guo, Jiang, Wang, Shi, Wang, and
  Li}]{shi2020pv}
Shi S, Guo C, Jiang L, Wang Z, Shi J, Wang X, Li H (2020{\natexlab{a}})
  Pv-rcnn: Point-voxel feature set abstraction for 3d object detection. In:
  Proceedings of the IEEE/CVF Conference on Computer Vision and Pattern
  Recognition, pp 10529--10538

\bibitem[{Shi et~al.(2020{\natexlab{b}})Shi, Wang, Shi, Wang, and
  Li}]{shi2020points}
Shi S, Wang Z, Shi J, Wang X, Li H (2020{\natexlab{b}}) From points to parts:
  3d object detection from point cloud with part-aware and part-aggregation
  network. IEEE Transactions on Pattern Analysis and Machine Intelligence
  43(8):2647--2664

\bibitem[{Sun et~al.(2020{\natexlab{a}})Sun, Cao, Chen, and
  Mao}]{sun2020towards}
Sun J, Cao Y, Chen QA, Mao ZM (2020{\natexlab{a}}) Towards robust
  $\{$LiDAR-based$\}$ perception in autonomous driving: General black-box
  adversarial sensor attack and countermeasures. In: 29th USENIX Security
  Symposium (USENIX Security 20), pp 877--894

\bibitem[{Sun et~al.(2021)Sun, Cao, Choy, Yu, Anandkumar, Mao, and
  Xiao}]{sun2021adversarially}
Sun J, Cao Y, Choy CB, Yu Z, Anandkumar A, Mao ZM, Xiao C (2021) Adversarially
  robust 3d point cloud recognition using self-supervisions. Advances in Neural
  Information Processing Systems 34:15498--15512

\bibitem[{Sun et~al.(2020{\natexlab{b}})Sun, Kretzschmar, Dotiwalla, Chouard,
  Patnaik, Tsui, Guo, Zhou, Chai, Caine, Vasudevan, Han, Ngiam, Zhao, Timofeev,
  Ettinger, Krivokon, Gao, Joshi, Zhang, Shlens, Chen, and
  Anguelov}]{Sun_2020_CVPR}
Sun P, Kretzschmar H, Dotiwalla X, Chouard A, Patnaik V, Tsui P, Guo J, Zhou Y,
  Chai Y, Caine B, Vasudevan V, Han W, Ngiam J, Zhao H, Timofeev A, Ettinger S,
  Krivokon M, Gao A, Joshi A, Zhang Y, Shlens J, Chen Z, Anguelov D
  (2020{\natexlab{b}}) Scalability in perception for autonomous driving: Waymo
  open dataset. In: Proceedings of the IEEE/CVF Conference on Computer Vision
  and Pattern Recognition, pp 2443--2451

\bibitem[{Szegedy et~al.(2013)Szegedy, Zaremba, Sutskever, Bruna, Erhan,
  Goodfellow, and Fergus}]{szegedy2013intriguing}
Szegedy C, Zaremba W, Sutskever I, Bruna J, Erhan D, Goodfellow I, Fergus R
  (2013) Intriguing properties of neural networks. International Conference on
  Learning Representations pp 1--10

\bibitem[{Tram{\`e}r et~al.(2018)Tram{\`e}r, Kurakin, Papernot, Goodfellow,
  Boneh, and McDaniel}]{tramer2018ensemble}
Tram{\`e}r F, Kurakin A, Papernot N, Goodfellow I, Boneh D, McDaniel P (2018)
  Ensemble adversarial training: Attacks and defenses. In: International
  Conference on Learning Representations

\bibitem[{Tsai et~al.(2020)Tsai, Yang, Ho, and Jin}]{tsai2020robust}
Tsai T, Yang K, Ho TY, Jin Y (2020) Robust adversarial objects against deep
  learning models. In: Proceedings of the AAAI Conference on Artificial
  Intelligence, vol~34, pp 954--962

\bibitem[{Tu et~al.(2020)Tu, Ren, Manivasagam, Liang, Yang, Du, Cheng, and
  Urtasun}]{TuRMLYDCU20}
Tu J, Ren M, Manivasagam S, Liang M, Yang B, Du R, Cheng F, Urtasun R (2020)
  Physically realizable adversarial examples for lidar object detection. In:
  Proceedings of the IEEE/CVF conference on computer vision and pattern
  recognition, Computer Vision Foundation / {IEEE}, pp 13713--13722

\bibitem[{Tu et~al.(2021)Tu, Li, Yan, Ren, Chen, Liang, Bitar, Yumer, and
  Urtasun}]{tu2021exploring}
Tu J, Li H, Yan X, Ren M, Chen Y, Liang M, Bitar E, Yumer E, Urtasun R (2021)
  Exploring adversarial robustness of multi-sensor perception systems in self
  driving. In: 5th Annual Conference on Robot Learning, pp 1--12

\bibitem[{Wen et~al.(2020)Wen, Lin, Chen, Chen, and Jia}]{wen2020geometry}
Wen Y, Lin J, Chen K, Chen CP, Jia K (2020) Geometry-aware generation of
  adversarial point clouds. IEEE Transactions on Pattern Analysis and Machine
  Intelligence

\bibitem[{Wicker and Kwiatkowska(2019)}]{wicker2019robustness}
Wicker M, Kwiatkowska M (2019) Robustness of 3d deep learning in an adversarial
  setting. In: Proceedings of the IEEE/CVF Conference on Computer Vision and
  Pattern Recognition, pp 11767--11775

\bibitem[{Xiang et~al.(2019)Xiang, Qi, and Li}]{xiang2019generating}
Xiang C, Qi CR, Li B (2019) Generating 3d adversarial point clouds. In:
  Proceedings of the IEEE/CVF Conference on Computer Vision and Pattern
  Recognition, pp 9136--9144

\bibitem[{Xu et~al.(2017)Xu, Evans, and Qi}]{xu2017feature}
Xu W, Evans D, Qi Y (2017) Feature squeezing: Detecting adversarial examples in
  deep neural networks. arXiv preprint arXiv:170401155

\bibitem[{Yan et~al.(2018)Yan, Mao, and Li}]{yan2018second}
Yan Y, Mao Y, Li B (2018) Second: Sparsely embedded convolutional detection.
  Sensors 18(10):3337

\bibitem[{Yang et~al.(2020)Yang, Sun, Liu, and Jia}]{yang20203dssd}
Yang Z, Sun Y, Liu S, Jia J (2020) 3dssd: Point-based 3d single stage object
  detector. In: Proceedings of the IEEE/CVF Conference on Computer Vision and
  Pattern Recognition, pp 11040--11048

\bibitem[{Zhang et~al.(2022)Zhang, Hu, Xu, Ma, Wan, and Guo}]{zhang2022not}
Zhang Y, Hu Q, Xu G, Ma Y, Wan J, Guo Y (2022) Not all points are equal:
  Learning highly efficient point-based detectors for 3d lidar point clouds.
  In: Proceedings of the IEEE/CVF Conference on Computer Vision and Pattern
  Recognition, pp 18953--18962

\bibitem[{Zhang et~al.(2023{\natexlab{a}})Zhang, Zhang, Hou, Yuan, and
  Xing}]{zhang2023upidet}
Zhang Y, Zhang Q, Hou J, Yuan Y, Xing G (2023{\natexlab{a}}) Unleash the
  potential of image branch for cross-modal 3d object detection. In:
  Thirty-seventh Conference on Neural Information Processing Systems

\bibitem[{Zhang et~al.(2023{\natexlab{b}})Zhang, Zhang, Zhu, Hou, and
  Yuan}]{zhang2023glenet}
Zhang Y, Zhang Q, Zhu Z, Hou J, Yuan Y (2023{\natexlab{b}}) Glenet: Boosting 3d
  object detectors with generative label uncertainty estimation. International
  Journal of Computer Vision pp 1--21

\bibitem[{Zhang et~al.(2023{\natexlab{c}})Zhang, Zhu, and
  Hou}]{zhang2023spatial}
Zhang Y, Zhu Z, Hou J (2023{\natexlab{c}}) Spatial-temporal enhanced
  transformer towards multi-frame 3d object detection. arXiv preprint
  arXiv:230700347

\bibitem[{Zheng et~al.(2019)Zheng, Chen, Yuan, Li, and
  Ren}]{zheng2019pointcloud}
Zheng T, Chen C, Yuan J, Li B, Ren K (2019) Pointcloud saliency maps. In:
  Proceedings of the IEEE/CVF International Conference on Computer Vision, pp
  1598--1606

\bibitem[{Zhou et~al.(2019)Zhou, Chen, Zhang, Fang, Zhou, and Yu}]{zhou2019dup}
Zhou H, Chen K, Zhang W, Fang H, Zhou W, Yu N (2019) Dup-net: Denoiser and
  upsampler network for 3d adversarial point clouds defense. In: Proceedings of
  the IEEE/CVF International Conference on Computer Vision, pp 1961--1970

\bibitem[{Zhou and Tuzel(2018)}]{Zhou_2018_CVPR}
Zhou Y, Tuzel O (2018) Voxelnet: End-to-end learning for point cloud based 3d
  object detection. In: Proceedings of the IEEE/CVF Conference on Computer
  Vision and Pattern Recognition, pp 4490--4499

\end{thebibliography}

\clearpage

\section*{Appendix}

The appendix offers additional experimental results not covered in the main content. Section \textcolor{red}{A} presents quantitative results of detectors under adversarial attacks. In Section \textcolor{red}{B}, we provide qualitative illustrations of the adversarial attacks and the defense strategies we studied.

\section*{A. Result of Absolute mAP Under Attacks}
In addition to the mAP ratio of detectors under adversarial attacks, we also report the absolute mAP values from Table~\ref{table:appendix_kitti_first} to Table~\ref{table:appendix_waymo_last}. As shown in Table~\ref{table:appendix_kitti_pgd}, the PDV achieves the best detection accuracy on the clean input but sub-optimal performance under the PGD-based point perturbation attack. The result implies that we should also take the robustness of 3D detectors into consideration in addition to detection accuracy.

\setlength{\tabcolsep}{4pt}
\begin{table}[htp]
	\centering
	\caption{The absolute mAP of various 3D detectors on the KITTI dataset under FGM-based point perturbation attacks with different $\epsilon$ values. $\epsilon$=0 denotes the mAP on clean input.}
	\label{table:appendix_kitti_first}
	\begin{tabular}{c|ccccccc} 
		\toprule
		\multirow{2}{*}{Detector} & \multicolumn{7}{c}{$\epsilon$ (cm)}  \\
		& 0     & 0.5 & 1  & 3  & 5  & 7  & 10    \\ 
		\midrule
		SECOND                     & 69.64 & 69.00 & 68.29 & 67.39 & 65.56 & 63.61 & 57.82  \\
		Voxel-RCNN                 & 74.36 & 73.88 & 73.68 & 74.17 & 72.45 & 71.13 & 67.21  \\
		PDV                        & 75.91 & 74.97 & 73.71 & 70.14 & 67.15 & 57.99 & 45.05  \\
		PointPillar                & 67.22 & 59.96 & 56.56 & 45.98 & 42.66 & 41.08 & 40.51  \\
		Part-A\^{}2                & 74.85 & 72.80 & 74.33 & 73.07 & 72.98 & 71.18 & 67.10  \\
		PV-RCNN                    & 71.03 & 71.03 & 70.27 & 67.67 & 65.99 & 61.50 & 51.24  \\
		3DSSD                      & 71.48 & 67.41 & 66.08 & 58.11 & 56.38 & 49.70 & 42.21  \\
		PointRCNN                  & 72.62 & 67.44 & 63.95 & 56.62 & 50.89 & 44.27 & 32.97  \\
		\bottomrule
	\end{tabular}
\end{table}
\vspace{-0.8cm}

\begin{table}[htp]
	\centering
	\caption{The absolute mAP of various 3D detectors on the KITTI dataset under MI-FGM-based point perturbation attacks with different $\epsilon$ values.}
	\begin{tabular}{c|ccccccc} 
		\toprule
		\multirow{2}{*}{Detector} & \multicolumn{7}{c}{$\epsilon$ (cm)}  \\
		& 0     & 0.5 & 1  & 3  & 5  & 7  & 10    \\ 
		\midrule
		SECOND                     & 69.64 & 68.91 & 68.24 & 67.37 & 65.63 & 63.94 & 57.38  \\
		Voxel-RCNN                 & 74.36 & 74.03 & 74.22 & 73.65 & 72.54 & 70.48 & 68.56  \\
		PDV                        & 75.91 & 73.80 & 71.88 & 66.78 & 58.35 & 49.01 & 37.14  \\
		PointPillar                & 67.22 & 60.30 & 54.88 & 39.38 & 32.07 & 28.59 & 26.58  \\
		Part-A\^{}2                & 74.85 & 73.03 & 73.21 & 73.15 & 73.78 & 70.56 & 67.96  \\
		PV-RCNN                    & 71.03 & 70.58 & 70.21 & 66.09 & 60.78 & 50.72 & 32.68  \\
		3DSSD                      & 71.48 & 66.26 & 63.25 & 49.98 & 39.71 & 26.71 & 15.92  \\
		PointRCNN                  & 72.62 & 65.13 & 62.60 & 48.43 & 34.14 & 25.29 & 12.90  \\
		\bottomrule
	\end{tabular}
\end{table}
\vspace{-1.3cm}

\begin{table}[htp]
	\centering
	\caption{The absolute mAP of various 3D detectors on the KITTI dataset under PGD-based point perturbation attacks with different $\epsilon$ values.}
	\label{table:appendix_kitti_pgd}
	\begin{tabular}{c|ccccccc} 
		\toprule
		\multirow{2}{*}{Detector} & \multicolumn{7}{c}{$\epsilon$ (cm)}  \\
		& 0     & 0.5 & 1  & 3  & 5  & 7  & 10    \\ 
		\midrule
		SECOND                     & 69.64 & 69.34 & 68.01 & 66.24 & 64.03 & 59.71 & 49.30  \\
		Voxel-RCNN                 & 74.36 & 73.14 & 74.01 & 72.29 & 70.81 & 68.13 & 59.67  \\
		PDV                        & 75.91 & 75.03 & 74.13 & 72.10 & 67.10 & 62.02 & 46.77  \\
		PointPillar                & 67.22 & 61.03 & 54.78 & 35.75 & 22.09 & 12.66 & 4.90   \\
		Part-A\^{}2                & 74.85 & 73.28 & 73.29 & 73.01 & 71.80 & 68.28 & 56.22  \\
		PV-RCNN                    & 71.03 & 70.89 & 68.36 & 66.43 & 63.04 & 54.39 & 35.37  \\
		3DSSD                      & 71.48 & 68.78 & 63.47 & 54.96 & 40.56 & 26.77 & 13.63  \\
		PointRCNN                  & 72.62 & 67.03 & 65.46 & 52.52 & 39.88 & 25.68 & 10.49  \\
		\bottomrule
	\end{tabular}
\end{table}
\vspace{-0.6cm}

\begin{table}[htp]
	\centering
	\caption{The absolute mAP of various 3D detectors on the KITTI dataset under point detachment attacks with different ratios of dropped points.}
	\begin{tabular}{c|ccccccc} 
		\toprule
		\multirow{2}{*}{Detector} & \multicolumn{7}{c}{Ratio of dropped points (\%)}  \\
		& 0     & 0.05 & 0.1  & 0.3  & 0.5  & 1  & 2    \\ 
		\midrule
        SECOND                     & 69.64 & 68.60 & 64.83 & 62.92 & 58.31 & 54.31 & 47.30  \\
        Voxel-RCNN                 & 74.36 & 74.02 & 71.06 & 65.64 & 62.62 & 58.18 & 48.21  \\
        PDV                        & 75.91 & 73.36 & 68.46 & 62.10 & 54.57 & 49.11 & 36.54  \\
        PointPillar                & 67.22 & 65.70 & 63.39 & 58.79 & 55.67 & 51.92 & 43.79  \\
        Part-A\^{}2                & 74.85 & 74.85 & 72.15 & 67.29 & 63.67 & 58.98 & 48.31  \\
        PV-RCNN                    & 71.03 & 70.44 & 68.87 & 64.59 & 62.40 & 56.48 & 44.90  \\
        3DSSD                      & 71.48 & 69.84 & 66.16 & 60.33 & 59.21 & 50.81 & 41.28  \\
        PointRCNN                  & 72.62 & 71.61 & 66.66 & 63.07 & 61.57 & 57.47 & 49.74  \\
		\bottomrule
	\end{tabular}
\end{table}

\begin{table}[htp]
	\centering
	\caption{The absolute mAP of various 3D detectors on the KITTI dataset under point attachment attacks with different $\epsilon$ values. The number of added points is set to 164.}
	\begin{tabular}{c|ccccccc} 
		\toprule
		\multirow{2}{*}{Detector} & \multicolumn{7}{c}{$\epsilon$ (m)}  \\
		& 0     & 0.05 & 0.1  & 0.3  & 0.5  & 0.7  & 1    \\ 
		\midrule
		SECOND                     & 69.64 & 66.03 & 64.72 & 59.54 & 53.59 & 58.15 & 60.40  \\
		Voxel-RCNN                 & 74.36 & 74.24 & 73.07 & 68.01 & 62.28 & 63.82 & 66.82  \\
		PDV                        & 75.91 & 72.23 & 67.11 & 48.69 & 48.29 & 49.98 & 58.49  \\
		PointPillar                & 67.22 & 65.45 & 64.21 & 60.37 & 57.80 & 55.40 & 55.39  \\
		Part-A\^{}2                & 74.85 & 73.13 & 73.56 & 69.30 & 68.47 & 70.53 & 71.47  \\
		PV-RCNN                    & 71.03 & 70.98 & 68.70 & 64.32 & 63.05 & 62.53 & 62.52  \\
		3DSSD                      & 71.48 & 49.28 & 45.29 & 48.80 & 52.81 & 53.91 & 59.23  \\
		PointRCNN                  & 72.62 & 69.42 & 63.55 & 52.91 & 52.85 & 53.87 & 56.80  \\
		\bottomrule
	\end{tabular}
\end{table}

\clearpage

\begin{table}[htp]
	\centering
	\caption{The absolute mAP of various 3D detectors on the Waymo dataset under FGM-based point perturbation attacks with different $\epsilon$ values. }
	\begin{tabular}{c|ccccccc} 
		\toprule
		\multirow{2}{*}{Detector} & \multicolumn{7}{c}{$\epsilon$ (cm)}  \\
		& 0     & 0.5 & 1  & 3  & 5  & 7  & 10    \\ 
		\midrule
			SECOND                    & 59.342 & 59.12 & 57.85 & 56.00 & 53.44 & 52.31 & 48.38  \\
			Voxel-RCNN                & 63.245 & 63.06 & 63.11 & 62.74 & 61.38 & 60.05 & 56.73  \\
			PDV                       & 68.640 & 68.43 & 67.57 & 67.25 & 66.58 & 65.75 & 62.13  \\
			PointPillar               & 57.016 & 52.57 & 49.70 & 40.23 & 34.38 & 31.07 & 27.87  \\
		    Part-A\^{}2               & 65.719 & 65.70 & 65.06 & 64.88 & 63.88 & 62.19 & 59.23  \\
			PV-RCNN                   & 65.603 & 64.68 & 64.58 & 62.55 & 61.17 & 58.17 & 54.47  \\
			3DSSD                     & 60.127 & 57.52 & 55.84 & 50.26 & 45.37 & 39.80 & 33.80  \\
			PointRCNN                 & 61.870 & 57.96 & 55.39 & 49.59 & 42.24 & 35.45 & 26.72  \\
		\bottomrule
	\end{tabular}
\end{table}

\begin{table}[htp]
	\centering
	\caption{The absolute mAP of various 3D detectors on the Waymo dataset under MI-FGM-based point perturbation attacks with different $\epsilon$ values. }
	\begin{tabular}{c|ccccccc} 
		\toprule
		\multirow{2}{*}{Detector} & \multicolumn{7}{c}{$\epsilon$ (cm)}  \\
		& 0     & 0.5 & 1  & 3  & 5  & 7  & 10    \\ 
		\midrule
        SECOND                    & 59.342 & 59.08 & 57.92 & 55.87 & 53.46 & 52.41 & 48.75  \\
        Voxel-RCNN                & 63.245 & 62.95 & 62.94 & 62.19 & 60.99 & 58.17 & 54.52  \\
        PDV                       & 68.640 & 68.41 & 67.72 & 67.24 & 65.92 & 63.78 & 63.03   \\
        PointPillar               & 57.016 & 52.57 & 48.14 & 35.02 & 25.24 & 19.11 & 13.11  \\
        Part-A\^{}2                  & 65.719 & 65.65 & 65.59 & 65.08 & 63.85 & 62.67 & 59.16  \\
        PV-RCNN                   & 65.603 & 64.51 & 64.00 & 60.58 & 58.20 & 53.47 & 46.08  \\
        3DSSD                     & 60.127 & 57.02 & 53.51 & 43.55 & 31.91 & 20.82 & 13.74  \\
        PointRCNN                 & 61.870 & 56.89 & 54.20 & 41.84 & 28.29 & 19.27 & 9.45   \\
		\bottomrule
	\end{tabular}
\end{table}

\begin{table}[htp]
	\centering
	\caption{The absolute mAP of various 3D detectors on the Waymo dataset under PGD-based point perturbation attacks with different $\epsilon$ values.}
	\begin{tabular}{c|ccccccc} 
		\toprule
		\multirow{2}{*}{Detector} & \multicolumn{7}{c}{$\epsilon$ (cm)}  \\
		& 0     & 0.5 & 1  & 3  & 5  & 7  & 10    \\ 
		\midrule
        SECOND                    & 59.342 & 59.16 & 58.11 & 55.88 & 52.40 & 49.48 & 42.92  \\
        Voxel-RCNN                & 63.245 & 63.06 & 62.80 & 62.23 & 62.08 & 59.86 & 56.74  \\
        PDV                       & 68.640 & 68.32 & 67.64 & 66.62 & 65.14 & 62.23 & 57.91  \\
        PointPillar               & 57.016 & 53.42 & 49.53 & 34.53 & 18.34 & 8.95  & 2.82   \\
        Part-A\^{}2                  & 65.719 & 65.77 & 65.70 & 64.61 & 63.30 & 62.83 & 58.66  \\
        PV-RCNN                   & 65.603 & 64.40 & 63.87 & 61.58 & 59.05 & 54.93 & 47.50  \\
        3DSSD                     & 60.127 & 58.58 & 54.03 & 47.96 & 32.07 & 20.83 & 9.81   \\
        PointRCNN                 & 61.870 & 57.73 & 56.57 & 46.38 & 33.03 & 19.27 & 7.50   \\
		\bottomrule
	\end{tabular}
\end{table}

\begin{table}[htp]
	\centering
	\caption{The absolute mAP of various 3D detectors on the Waymo dataset under point detachment attacks with different ratios of dropped points.}
	\begin{tabular}{c|ccccccc} 
		\toprule
		\multirow{2}{*}{Detector} & \multicolumn{7}{c}{Ratio of dropped points (\%)}  \\
		& 0     & 0.05 & 0.1  & 0.3  & 0.5  & 1  & 2    \\ 
		\midrule
        SECOND                    & 59.342 & 58.26 & 55.99 & 53.62 & 51.95 & 50.31 & 45.06  \\
        Voxel-RCNN                & 63.245 & 60.62 & 57.64 & 54.22 & 51.78 & 48.40 & 41.02  \\
        PDV                       & 68.640 & 66.00 & 62.54 & 58.80 & 56.76 & 52.97 & 46.14  \\
        PointPillar               & 57.016 & 56.55 & 54.79 & 53.13 & 52.20 & 50.86 & 46.84  \\
        Part-A\^{}2                  & 65.719 & 63.23 & 59.64 & 56.72 & 53.99 & 51.73 & 45.26  \\
        PV-RCNN                   & 65.603 & 61.97 & 56.61 & 50.31 & 46.16 & 41.26 & 29.43  \\
        3DSSD                     & 60.127 & 56.52 & 52.99 & 49.22 & 48.40 & 41.71 & 33.54  \\
        PointRCNN                 & 61.870 & 58.30 & 53.78 & 52.19 & 50.35 & 48.00 & 39.73  \\
		\bottomrule
	\end{tabular}
\end{table}

\begin{table}[htp]
	\centering
	\caption{The absolute mAP of various 3D detectors on the Waymo dataset under point attachment attacks with different $\epsilon$ values. $\epsilon$=0 denotes the mAP on clean input. The number of added points is set to 500.}
	\label{table:appendix_waymo_last}
	\begin{tabular}{c|ccccccc} 
		\toprule
		\multirow{2}{*}{Detector} & \multicolumn{7}{c}{$\epsilon$ (m)}  \\
		& 0     & 0.05 & 0.1  & 0.3  & 0.5  & 0.7  & 1    \\ 
		\midrule
        SECOND                    & 59.342 & 58.26 & 56.53 & 50.53 & 46.25 & 44.56 & 45.61  \\
        Voxel-RCNN                & 63.245 & 54.66 & 54.09 & 57.21 & 58.95 & 61.16 & 62.23  \\
        PDV                       & 68.640 & 67.82 & 67.76 & 62.71 & 58.98 & 57.61 & 57.37   \\
        PointPillar               & 57.016 & 55.93 & 55.45 & 51.53 & 48.89 & 46.81 & 46.27  \\
        Part-A\^{}2                  & 65.719 & 65.65 & 64.58 & 59.75 & 54.76 & 53.75 & 54.60  \\
        PV-RCNN                   & 65.603 & 60.01 & 59.32 & 60.06 & 60.26 & 61.85 & 62.56  \\
        3DSSD                     & 60.127 & 40.58 & 37.32 & 41.64 & 46.28 & 49.11 & 54.24  \\
        PointRCNN                 & 61.870 & 54.41 & 50.73 & 45.75 & 45.94 & 48.26 & 51.20  \\
		\bottomrule
	\end{tabular}
\end{table}

\section*{B. Visual Results}\label{sec:appendix_visual_results}

We present a qualitative analysis that helps us better understand the adversarial attacks against 3D object detectors and the effectiveness of defense.

\begin{figure*}[htp]
	\centering
	\includegraphics[width=0.99\textwidth]{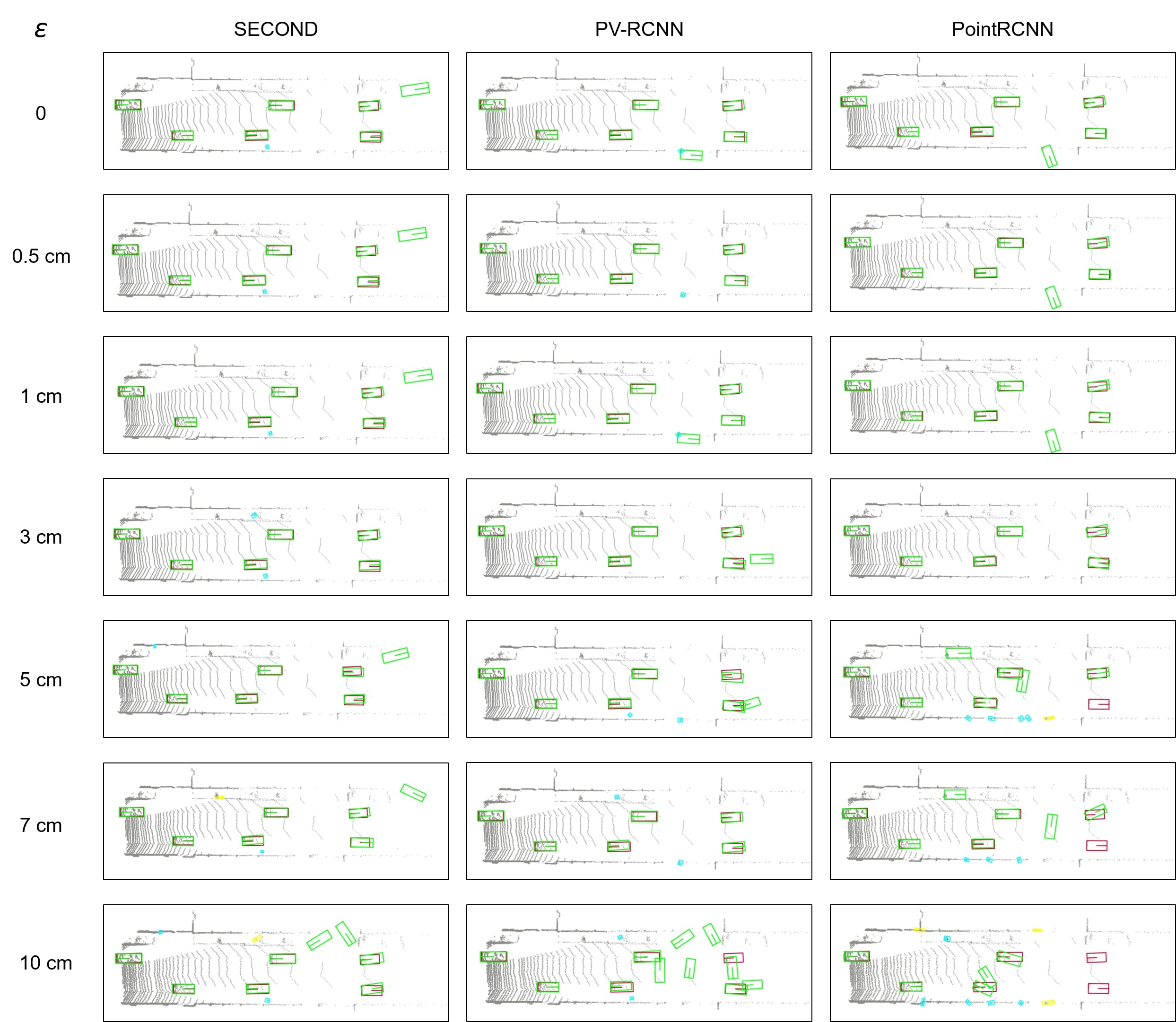} 
	\caption{
			Visual results of three detectors when PGD-based point perturbation attacks with various $\epsilon$ values are applied. The predicted bounding boxes of \textcolor{green}{car}, \textcolor{cyan}{pedestrian}, and \textcolor{yellow}{cyclist} are visualized in \textcolor{green}{green}, \textcolor{cyan}{cyan}, and \textcolor{yellow}{yellow}, respectively. The \textcolor{red}{ground-truth} bounding boxes are visualized in \textcolor{red}{red}. Best viewed in color and zoom in for more details.
		}
	\label{fig:vis_perturbation_on_different_models}
\end{figure*}

In Figure~\ref{fig:vis_perturbation_on_different_models}, we visually illustrate the detection results under PGD-based point perturbation attacks with varying $\epsilon$ values. As the $\epsilon$ value increases, indicating more pronounced perturbations, the performance of detectors tends to degrade. This decline is evidenced by more false-positive boxes and missed detections of objects. Besides, among the variety of detectors evaluated, the point-based ones, especially PointRCNN, seem to be more susceptible to this adversarial attack, revealing a crucial vulnerability.

\begin{figure*}[t]
	\centering
	\includegraphics[width=0.99\textwidth]{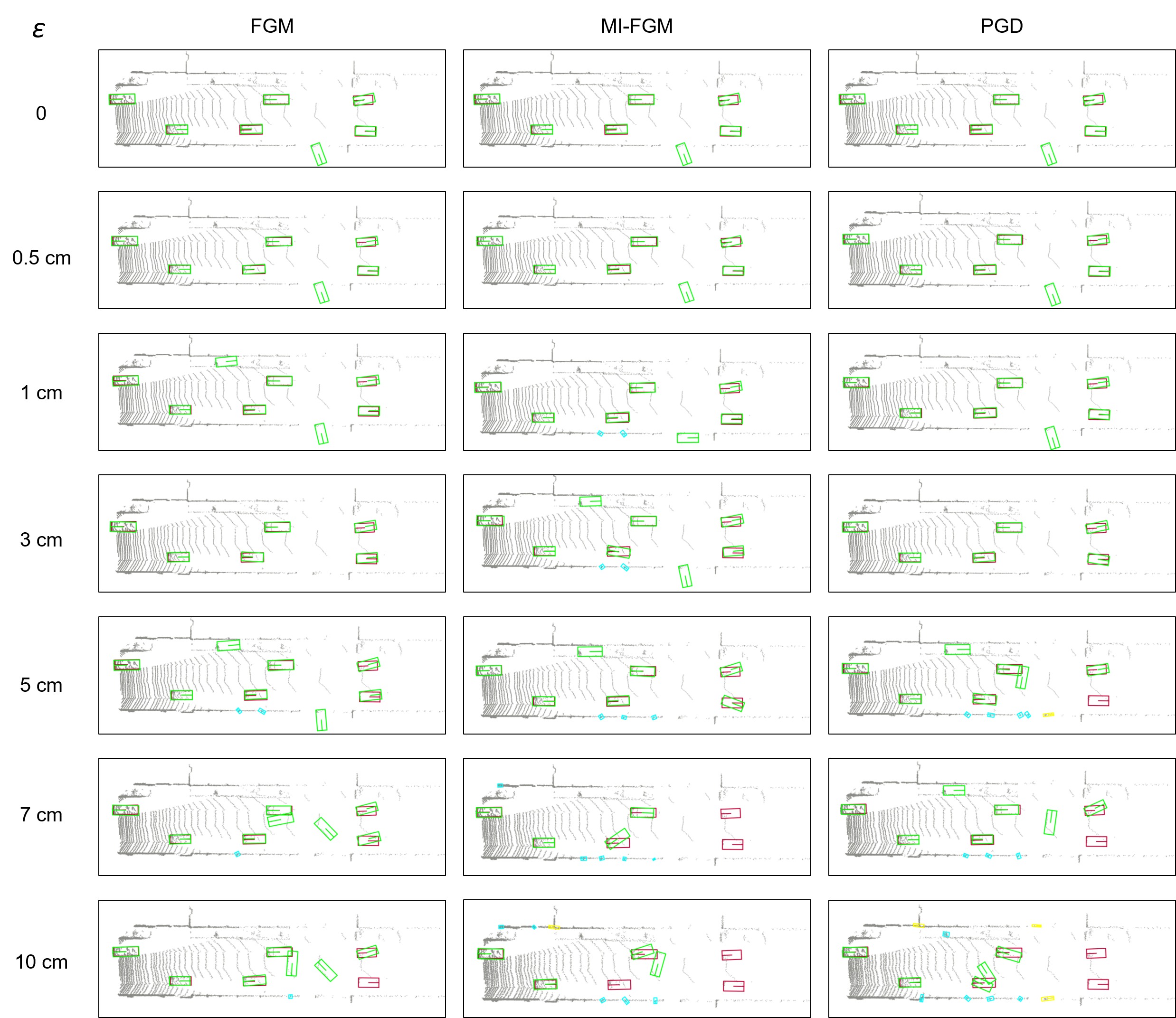} 
	\caption{
			Comparison of different types of point perturbation attacks applied on PointRCNN. The predicted bounding boxes of \textcolor{green}{car}, \textcolor{cyan}{pedestrian}, and \textcolor{yellow}{cyclist} are visualized in \textcolor{green}{green}, \textcolor{cyan}{cyan}, and \textcolor{yellow}{yellow}, respectively. The \textcolor{red}{ground-truth} bounding boxes are visualized in red. Best viewed in color and zoom in for more details.
		}
	\label{fig:vis_different_perturbation_attack}
\end{figure*}

Subsequently, Figure~\ref{fig:vis_different_perturbation_attack} offers a granular perspective on the adversarial robustness of PointRCNN under varied point perturbation attack modalities. The iterative techniques, particularly the MI-FGM and PGD-based attacks, exhibit pronounced adversarial strength compared to the more single-step FGM method. Among these, the PGD-based attack conspicuously overshadows the rest in its disruptive prowess. This visual deduction is congruent with our quantitative findings presented in Section~\ref{sec:evaluation_perturbation_attack}.

\begin{figure*}[t]
	\centering
	\includegraphics[width=0.85\textwidth]{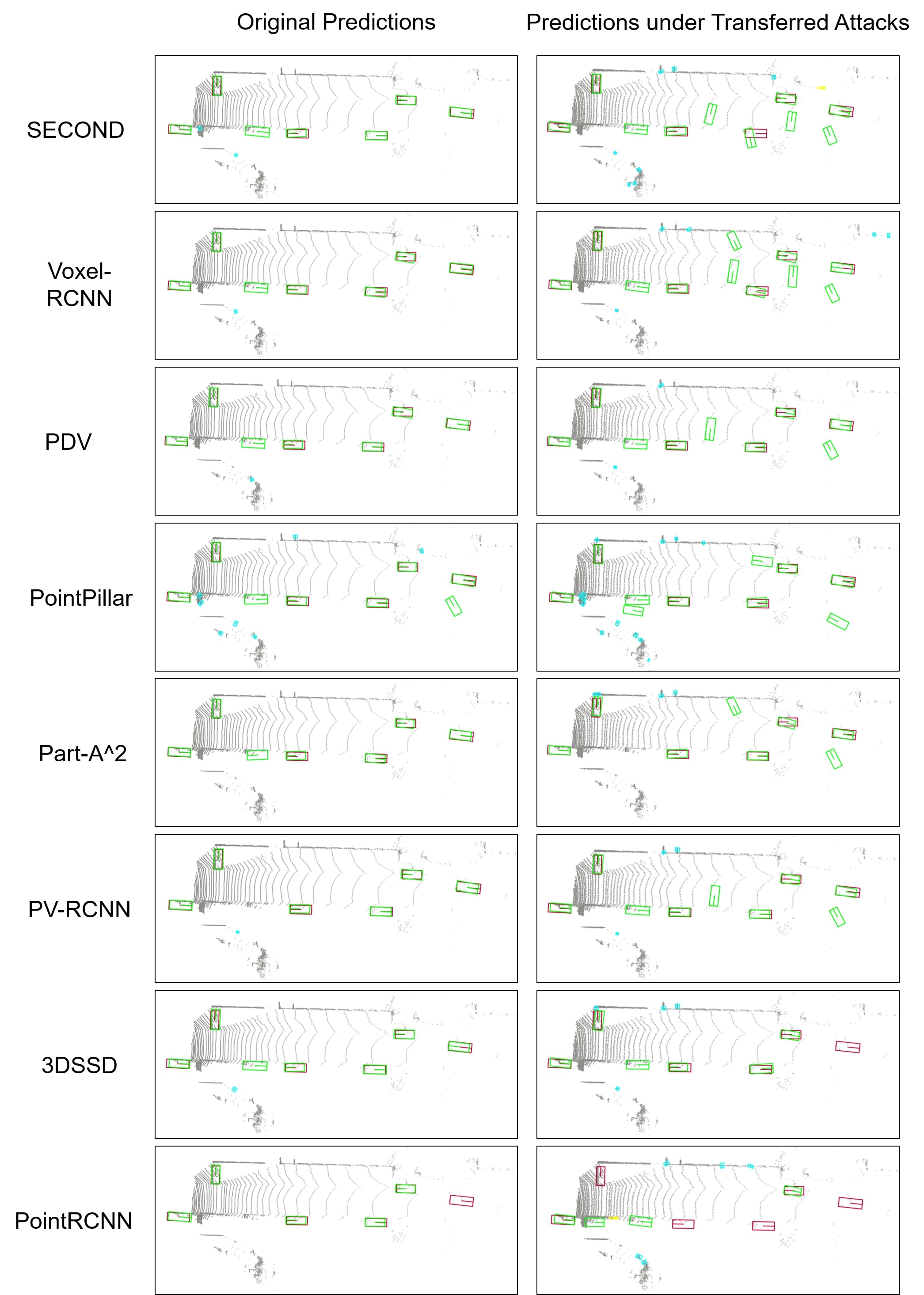} 
	\caption{
			Detection results of state-of-the-art detectors on clean input and the transferred adversarial examples generated from PointRCNN under point perturbation attack. The predicted bounding boxes of \textcolor{green}{car}, \textcolor{cyan}{pedestrian}, and \textcolor{yellow}{cyclist} are visualized in \textcolor{green}{green}, \textcolor{cyan}{cyan}, and \textcolor{yellow}{yellow}, respectively. The \textcolor{red}{ground-truth} bounding boxes are visualized in \textcolor{red}{red}. Best viewed in color and zoom in for more details.
		}
	\label{fig:vis_cross_model_perturbation_attack}
\end{figure*}

Turning to Figure~\ref{fig:vis_cross_model_perturbation_attack}, we broach the subject of attack transferability
It is interesting to see how an attack, originally crafted for PointRCNN, retains its disruptive capabilities when applied to other models. The visual results showcase increased false positives and overlooked objects, underscoring the wide-ranging effectiveness of such transferred attacks.

\begin{figure*}[t]
	\centering
	\includegraphics[width=\textwidth]{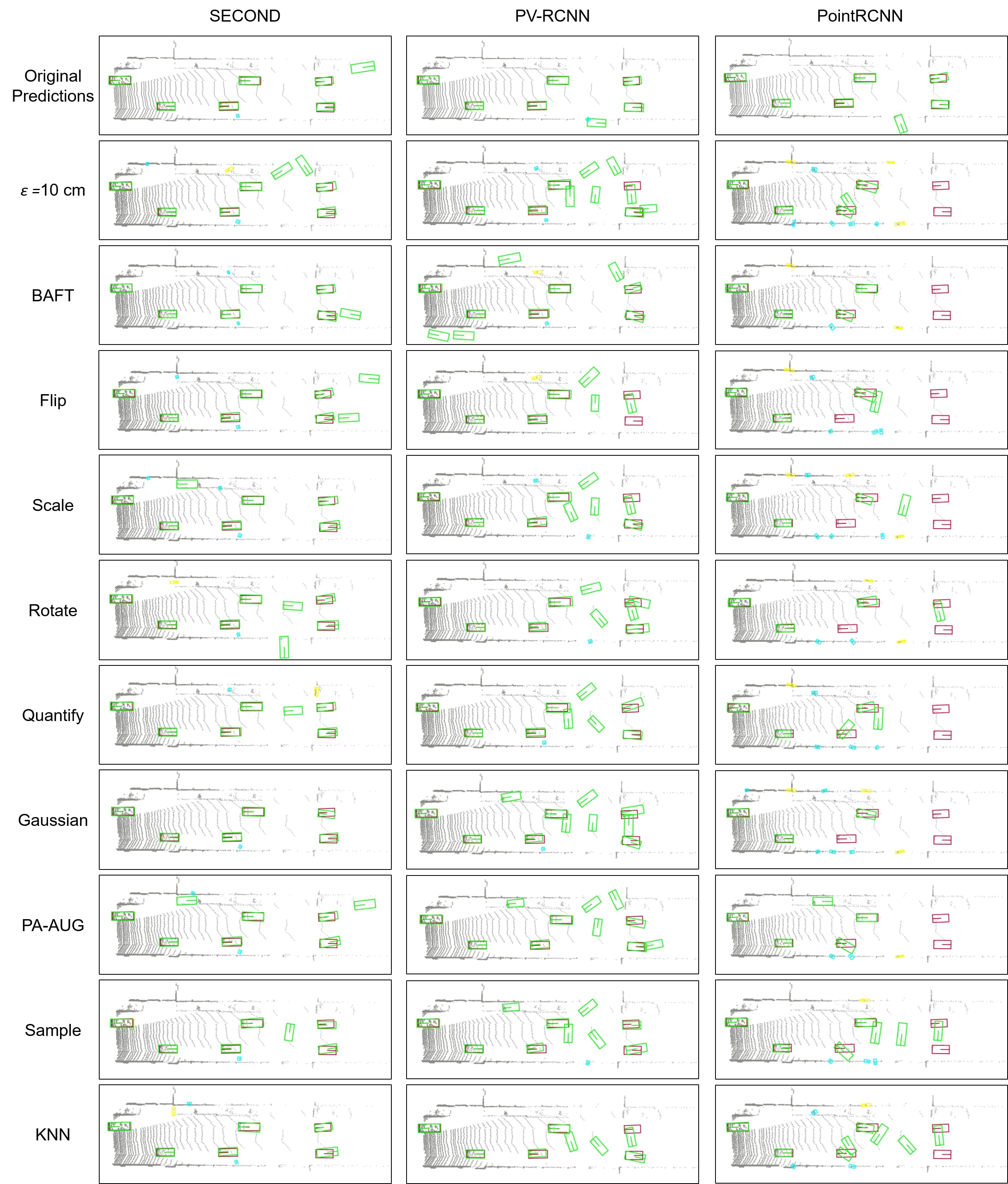} 
	\caption{
			Visual results of detectors on adversarial examples generated by point perturbation attack after applying various defenses. The predicted bounding boxes of \textcolor{green}{car}, \textcolor{cyan}{pedestrian}, and \textcolor{yellow}{cyclist} are visualized in \textcolor{green}{green}, \textcolor{cyan}{cyan}, and \textcolor{yellow}{yellow}, respectively. The \textcolor{red}{ground-truth} bounding boxes are visualized in \textcolor{red}{red}. Best viewed in color and zoom in for more details.
		}
	\label{fig:vis_different_defense_for_perturbation}
\end{figure*}

We also provide qualitative results in Figure~\ref{fig:vis_different_defense_for_perturbation}, we can clearly observe our proposed BAFT significantly improves the robustness of detectors under point perturbation attack. Besides, applying transformation-based defenses to the perturbed point clouds could slightly improve the robustness of detectors against adversarial attacks.

These visual representations not only offer a more intuitive understanding of the adversarial challenges but also highlight the specific vulnerabilities of prominent detection models.

\end{document}